%% file: paper.tex
\documentclass[11pt]{article}
\usepackage{etex}
\usepackage{acl2016}
\usepackage{times}
\usepackage{latexsym}
\usepackage{amsmath}
\usepackage{graphicx}
\usepackage{subcaption}
\usepackage{multirow}
\usepackage{adjustbox}
\usepackage{tikz-dependency}
\usepackage{url}
\usepackage{enumitem}
\usepackage{pstricks}
\usepackage{verbatim}
\usepackage{pgfplots}
\usepackage[font={small}]{caption}
\usepackage{booktabs}
\usepackage{tikz}
\usetikzlibrary{decorations.markings,arrows,positioning,fit,shapes,calc,chains,fit}
\usepackage{fancyref}
\usepackage{titlesec}
\titlespacing\section{0pt}{11pt plus 3pt minus 2pt}{1pt plus 2pt minus
2pt}
\titlespacing\subsection{0pt}{11pt plus 3pt minus 2pt}{0pt plus 2pt minus 2pt}
\titlespacing\subsubsection{0pt}{11pt plus 3pt minus 2pt}{0pt plus 2pt minus
2pt}
\makeatother
\edef\sizetape{0.7cm}
\tikzstyle{tmtape}=[draw,minimum size=\sizetape]
\tikzstyle{tmhead}=[arrow box,draw,minimum size=.5cm,arrow box
arrows={east:.25cm, west:0.25cm}]

\newcommand{\VOC}{\mathcal{V}}
\newcommand{\Tset}{\mathcal{T}}

\newcommand{\Dset}{\mathcal{D}}

\aclfinalcopy

\title{Mapping Unseen Words to Task-Trained Embedding Spaces}
\author{Pranava Swaroop Madhyastha$^{\ast}$
\ \ \ \  Mohit Bansal$^{\dag}$
\ \ \ \  Kevin Gimpel$^{\dag}$
\ \ \ \  Karen Livescu$^{\dag}$
\\
$^{\ast}$Universitat Polit\`ecnica de Catalunya\\
\tt{pranava@cs.upc.edu}\\
$^{\dag}$Toyota Technological Institute at Chicago\\
\tt{\{mbansal,kgimpel,klivescu\}@ttic.edu}
}

\date{}
\begin{document}
\maketitle
\begin{abstract}

We consider 
the supervised training 
setting in which we 
learn task-specific word embeddings. We assume 
that we start with 
initial 
embeddings learned from unlabelled data 
and update them to learn 
task-specific embeddings
for words 
in the 
supervised
training data.
However, for new words in the test set,   
we must use 
either their initial embeddings 
or a single unknown embedding, 
which often leads to errors. 
We address this 
by learning a neural network to map from initial 
embeddings 
to the task-specific embedding space, via a multi-loss objective function. 
The technique is general, but here we demonstrate its use for improved dependency
parsing (especially for sentences with
out-of-vocabulary words), as well as for downstream improvements on sentiment analysis.
\end{abstract}
\section{Introduction}
Performance on NLP tasks drops significantly when moving from training sets to held-out data~\cite{petrov2010uptraining}. 
One cause of this drop is 
words that do not appear in the 
training data but appear in 
test data, whether in the same domain or in a new domain. 
We refer to such out-of-training-vocabulary (OOTV) words as \emph{unseen}
words. 
NLP systems often make errors on unseen words and, in structured
tasks 
like dependency parsing, this can trigger  
a cascade of errors in the sentence. 

Word 
embeddings 
can 
counter the effects of limited training data~\cite{Necsulescuetal,turian2010word,collobert2011b}. 
While the 
effectiveness of 
pretrained embeddings can be 
heavily
task-dependent~\cite{bansal2014tailoring}, 
there is a great deal of work on
updating embeddings during supervised training to make them more 
task-specific~\cite{kalchbrenner-grefenstette-blunsom:2014:P14-1,qu2015big,chen2014fast}. 
These task-trained embeddings have shown encouraging results but raise some
concerns:  
(1) the updated embeddings of infrequent words are prone
to overfitting, and 
(2) many words in the test data are not contained in the training data
at all. 
In the latter case, at test time, 
systems either use 
a single, generic 
embedding for all unseen words or use their initial
embeddings (typically derived from unlabelled
data)~\cite{sgaardjohannsen2012POSTERS,collobert2011b}. 
Neither 
choice is ideal: 
A single unknown embedding 
conflates many words, 
while the
initial embeddings may be in a space that is not comparable to the
trained embedding space.

In this paper, we address both concerns by learning to 
map from the 
initial embedding space to the task-trained space. 
We train a neural network 
mapping function that takes initial word
embeddings 
and
maps them to task-specific 
embeddings that are trained for the given task, via a multi-loss objective
function. 
We tune the mapper's hyperparameters
to optimize performance on each domain of interest, thereby achieving some of
the benefits of domain adaptation. We demonstrate significant improvements in dependency parsing across several domains and for the downstream task of dependency-based sentiment analysis using the model of \newcite{tai2015improved}.

\section{Mapping Unseen Representations}
\label{sec:theory}

\begin{figure*}
\begin{subfigure}{.6\textwidth}
\centering
\adjustbox{width=1.2\linewidth}{\input{figs/learning.tex}}
\caption{Mapper Training}\label{fig:mapper} 
\label{fig:pipe1}
\end{subfigure}
\begin{subfigure}{.36\textwidth}
\centering
\adjustbox{width=1.2\linewidth}{\input{figs/parsing.tex}}
\caption{\small{Test-time: Parsing with Mapped Embeddings}}\label{fig:testing} 
\label{fig:pipe2}
\end{subfigure}\caption{System Pipeline} 
\label{fig:pipeline}
\end{figure*}
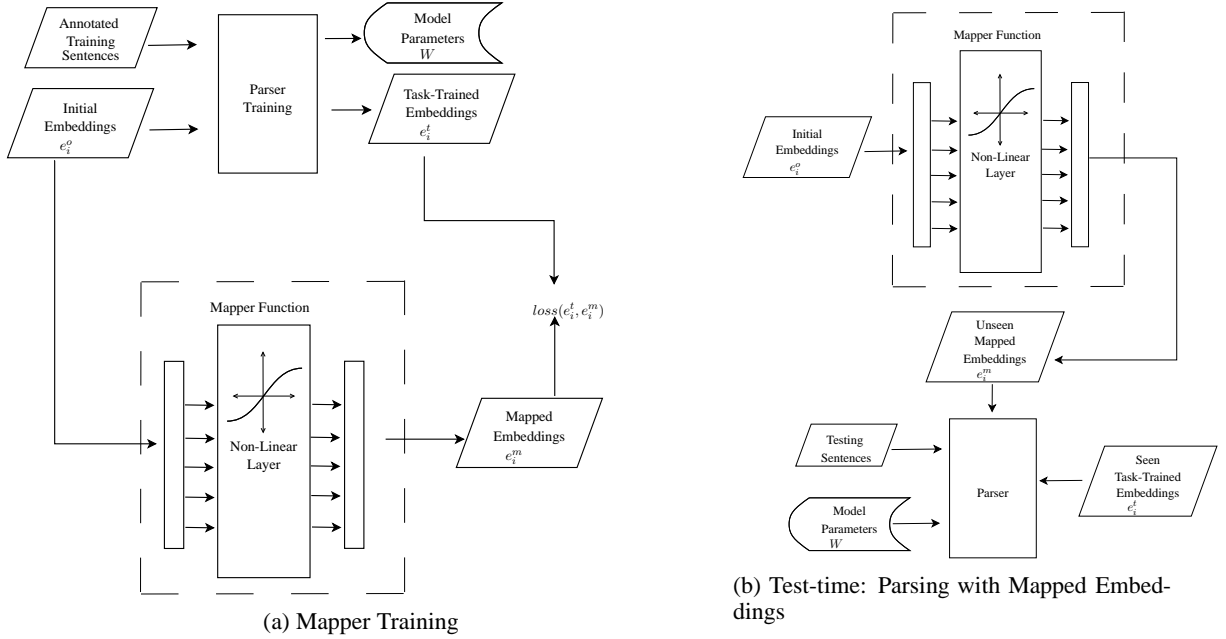

Let $\VOC = \{w_1, \ldots, w_V \}$ be the vocabulary of word types in a large, unannotated corpus. 
Let $e_{i}^{o}$ denote the initial (original) embedding of word $w_i$ computed from this
corpus.  
The initial embeddings are typically learned in an unsupervised way, but for our purposes they can be any initial embeddings.
Let $\Tset \subseteq \VOC$ be the subset of words that appear in the annotated training data for some supervised task-specific training. 
We define \textbf{unseen} words as those in the set $\VOC\setminus\Tset$. 
While our approach is general, for concreteness, we consider the task of dependency parsing, so the annotated data consists of sentences paired with dependency trees. 
We assume a dependency parser that learns task-specific word embeddings $e_{i}^{t}$ for word $w_i \in \Tset$, starting from the original embedding $e_{i}^{o}$. In this work, we use the Stanford neural dependency parser~\cite{chen2014fast}.

The goal of the mapper is as follows. We are given a training set of $N$ pairs of initial and task-trained embeddings
$\Dset=\left\{ \left( e_{1}^{o},e_{1}^{t}\right), \ldots, \left(e_{N}^{o},e_{N}^{t}\right)\right\}$,
and we want to learn a function $G$ that maps each initial embedding $e_{i}^{o}$ to be as close as possible to its corresponding output embedding $e_{i}^{t}$. 
We denote the mapped embedding $e_{i}^{m}$, i.e., $e_{i}^{m} = G\left(e_{i}^{o}\right)$.

Figure~\ref{fig:pipe1} describes the training procedure of the mapper. We use a supervised parser
which is trained on an annotated dataset and initialized with pre-trained word embeddings $e_{i}^{o}$. The parser uses back-propagation 
to update these embeddings during training, producing task-trained embeddings $e_{i}^{t}$ for all $w_i \in \Tset$.  
After we train the parser, the mapping function $G$ is trained to map an initial word 
embedding $e_{i}^{o}$ to its parser-trained embedding $e_{i}^{t}$. 
At test (or development) time, we use the trained mapper $G$ to transform the original embeddings of unseen test words to the parser-trained space (see Figure~\ref{fig:pipe2}).
When parsing held-out data, 
we use the same parser model parameters 
($W$) as shown in Figure~\ref{fig:pipe2}. 
The only difference is that now some of the word embeddings (i.e., for unseen words) have changed to mapped ones.

\subsection{Mapper Architecture}
Our proposed mapper is a multi-layer
feedforward neural network 
that takes an initial word embedding as input 
and outputs a mapped representation of the same dimensionality. 
In particular, we use a 
single hidden layer 
with a hard$\tanh$ non-linearity, so the function $G$ is defined as:
\begin{equation}
G(e_{i}^{o}) = W_2(\text{hard}\tanh(W_1e_{i}^{o} + b_1)) + b_2
\end{equation}
\noindent where $W_1$ and $W_2$ are parameter matrices and $b_1$ and $b_2$ are bias vectors.

The `hard$\tanh$' non-linearity is the standard `hard' version of hyperbolic tangent:
\begin{equation*}
    \text{hard}\tanh(z) = 
\begin{cases}
    {-1} & \text{if  } z < {-1}\\
    z & \text{if } {-1} \leq z \leq 1\\
    1 & \text{if  } z > 1
\end{cases}
\end{equation*}

In preliminary experiments we compared with other non-linear functions (sigmoid, $\tanh$, and ReLU), as well as with zero and more than one non-linear layers.
We found that 
fewer or more non-linear
layers did not improve performance. 
\subsection{Loss Function}
We use a weighted, multi-loss regression approach, optimizing a weighted sum of
mean squared error and mean absolute error:
\begin{align}
    &\mathit{loss}(y,\hat{y}) = \nonumber\\
&\alpha \sum\limits_{j=1}^{n}|y_j - \hat{y_j}| +(1-\alpha) \sum\limits_{j=1}^{n} |y_j - \hat{y_j}|^2\label{eq:loss}
\end{align}
\noindent where $y = e_{i}^{t}$ (the ground truth) and $\hat{y} = e_{i}^{m}$
(the prediction) are $n$-dimensional vectors. This multi-loss approach seeks to
make both the conditional \emph{mean} of the predicted representation close to
the task-trained representation (via the squared loss) and the conditional
\emph{median} of the predicted representation close to the task-trained one
(via the 
mean absolute loss). A weighted multi-criterion objective allows us to avoid making 
strong assumptions about the optimal transformation to be learned.  We tune the
hyperparameter $\alpha$ on domain-specific held-out data. 
We try to minimize the assumptions in our formulation of the loss, and let the
tuning determine the particular mapper configuration that works best for each
domain.
Strict squared loss or an absolute loss are just
special forms of this loss function.

For optimization, we use batch limited memory BFGS (L-BFGS)~\cite{liu89}. 
In preliminary experiments comparing with stochastic optimization, we found
L-BFGS to be more stable to train and easier to check for convergence (as has
recently been found in other settings as well~\cite{ngiam2011optimization}).

\subsection{Regularization} 
We use elastic net regularization~\cite{liu89}, which linearly combines $\ell_1$ and
$\ell_2$ penalties on the parameters to control the capacity of the mapper
function. This equates to minimizing:
\begin{equation*}
F(\theta) = L(\theta) + \lambda_1\|\theta\|_1 + \frac{\lambda_2}{2}\|\theta\|_2^2
\end{equation*}
\noindent where $\theta$ is the full set of mapper parameters and $L(\theta)$ is
the loss function (Eq.~\ref{eq:loss} summed over mapper training examples). We tune the hyperparameters of the regularizer
and the loss function separately for each task, using a task-specific development set.
This gives us additional flexibility to map the embeddings 
for the domain of interest, especially 
when the parser training data comes from a particular domain
(e.g., newswire) and we want to use the parser on a new domain
(e.g., email). 
We also tried dropout-based
regularization~\cite{srivastava14a}
for the non-linear layer but did not see any significant improvement.

\subsection{Mapper-Parser Thresholds}

Certain words in the parser training data $\Tset$ are very infrequent, 
which may lead to inferior task-specific embeddings $e_i^t$ 
learned by the parser. We want our mapper function to be learned on
high-quality task-trained embeddings. After learning a strong mapping function, we can
use it to remap the inferior task-trained embeddings.

We thus consider several frequency thresholds that control which word embeddings to use 
to train the mapper and which to map at test time. 
Below are the specific thresholds that we consider:
\paragraph{Mapper-training Threshold ($\tau_t$)} The mapper is trained only on
embedding pairs for words seen at least $\tau_t$ times in the training data $\Tset$. 
\paragraph{Mapping Threshold ($\tau_m$)} For test-time inference, the mapper will map 
any word whose count in $\Tset$ is less than $\tau_m$. That is, we discard parser-trained 
embeddings $e_i^t$ of these infrequent words and use our mapper to map the initial embeddings $e_i^o$ instead. 
\paragraph{Parser Threshold ($\tau_p$)} While training the parser, 
for words that appear fewer than $\tau_p$ times in $\Tset$, 
the parser replaces them with the 
`unknown' embedding. Thus, no parser-trained embeddings will be learned for these words. 

In our experiments, we explore a small set of values from this large space of possible threshold combinations (detailed below). 
We consider only relatively small values for the mapper-training ($\tau_t$) and parser thresholds ($\tau_p$) because as 
we increase them, the number of training examples for the mapper decreases,
making it harder to learn an accurate mapping function\footnote{Note that the
training of the
    mapper tends to be very quick because training examples are word types
    rather than word tokens. When we increase $\tau_t$, the number of training
    examples reduces further. Hence, since we do not have many examples, we
    want the mapping procedure to have as much flexibility as possible, so we
    use multiple losses and regularization strategies, and then tune their
relative strengths.}. 
\section{Related Work}
\label{sec:relwork}
There are several categories of related approaches, 
including those that 
learn a single embedding 
for unseen words~\cite{sgaardjohannsen2012POSTERS,chen2014fast,collobert2011b}, those that use character-level
information~\cite{luong-socher-manning:2013:CoNLL-2013,botha2014compositional,wang:2015,ballesteros2015improved}, 
those using morphological and $n$-gram information~\cite{candito2009improving,habash2009remoov,marton-habash-rambow:2010:SPMRL,seddah-EtAl:2010:SPMRL,attia2010handling,bansal2011web,keller2003using}, and 
hybrid approaches~\cite{dyer2015transition,jean-EtAl:2015:ACL-IJCNLP,luong2015addressing,chitnis-denero:2015:EMNLP}. 
The representation for the unknown token is either learned specifically
or computed from a selection of rare words, for example by averaging their embedding vectors.

Other work has also found improvements by combining pre-trained, fixed embeddings with
task-trained embeddings~\cite{kim:2014:EMNLP2014,paulus2014global}. 
Also relevant are approaches developed specifically to handle large target vocabularies (including many rare words) in neural 
machine translation systems~\cite{jean-EtAl:2015:ACL-IJCNLP,luong2015addressing,chitnis-denero:2015:EMNLP}. 

Closely related to our approach is that of \newcite{tafforeau2015alr}. 
They induce embeddings for unseen words by combining the embeddings of the $k$
nearest neighbors. In Sec.~\ref{sec:experiments}, we show that our approach
outperforms theirs. Also related is the approach taken by 
\newcite{kiros2015skip}. They learn a 
linear mapping of the initial embedding space 
via 
unregularized linear regression. 
Our approach differs by considering nonlinear mapping functions, comparing different losses and mapping thresholds, and  
learning 
separately tuned mappers for 
each domain of interest. 
Moreover, we focus on empirically evaluating the effect of the mapping of
unseen words, showing statistically significant improvements on both parsing
and a downstream task (sentiment analysis).

\section{Experimental Setup}

\label{sec:experiments}
\subsection{Dependency Parser}
We use the feed-forward neural network dependency parser
of~\newcite{chen2014fast}. In all
our experiments (unless stated otherwise), we use the default arc-standard
parsing configuration and hyperparameter settings. 
For evaluation, we compute the percentage of words that get the correct head,
reporting both unlabelled attachment score (UAS) and labelled attachment score (LAS). 
LAS additionally requires the predicted dependency label to be correct. 
To measure statistical significance, we use a bootstrap test~\cite{efron1986} with
100K samples. 
\subsection{Pre-Trained Word Embeddings}

We use the 100-dimensional GloVe word embeddings from
\newcite{pennington2014glove}. These were trained on Wikipedia 2014 and the
Gigaword v5 corpus and have a vocabulary size of approximately
400,000.\footnote{{\tiny{\url{http://www-nlp.stanford.edu/data/glove.6B.100d.txt.gz};}}
We have also
experimented with the downloadable 50-dimensional SENNA embeddings from
\newcite{collobert2011b} and with \texttt{word2vec}~\cite{mikolov2013efficient}
embeddings that we trained ourselves; in preliminary experiments
the GloVe embeddings performed best, so we use them for all experiments
below.}

\subsection{Datasets}

We consider a number of datasets with varying rates of OOTV words.  We define the OOTV rate (or, equivalently, the unseen rate) of a dataset as the percentage of the vocabulary (types) of words occurring in the set that were not seen in training.

\paragraph{Wall Street Journal (WSJ) and OntoNotes-WSJ}
We conduct experiments on the Wall Street Journal portion of the English Penn
Treebank dataset~\cite{marcus1993building}. We follow the standard splits: sections 2-21 for training, section 22 for
validation, and section 23 for testing.  We convert the original phrase
structure trees into dependency trees using Stanford Basic
Dependencies~\cite{de2008stanford} in the Stanford Dependency Parser. The POS
tags are obtained using the Stanford POS tagger~\cite{toutanova2003feature}
in a 
10-fold jackknifing setup on the training data (achieving an accuracy of
96.96\%). The OOTV rate in the development and test sets is approximately 2-3\%.

We also conduct experiments on the OntoNotes 4.0 dataset 
(which we denote OntoNotes-WSJ). This
dataset contains the same sentences as the WSJ corpus (and we use the same data splits), but has 
significantly different annotations. The OntoNotes-WSJ training data is used for 
the Web Treebank test experiments. We perform the same pre-processing steps as for the WSJ dataset. 

\paragraph{Web Treebank}
We expect our mapper to be most effective when parsing held-out data with many
unseen words. 
This often happens when the held-out data is drawn from a different
distribution than the training data. 
For example, when training a parser on newswire and testing on web data, 
many errors occur due to differing patterns of syntactic usage and unseen
words~\cite{foster2011hardtoparse,petrov2012overview,kong-EtAl:2014:EMNLP2014,wang-EtAl:2014:EMNLP20144}. 

We explore this setting by training our parser on OntoNotes-WSJ
and testing on the Web Treebank~\cite{petrov2012overview}, which includes five
domains: answers, email, newsgroups, reviews, and
weblogs. Each domain contains approximately 2000-4000 manually annotated
syntactic parse trees in the OntoNotes 4.0 style. 
In this case, we are adapting the parser which is trained on OntoNotes corpora
using the small development set for each of the sub-domains (the size of the
Web Treebank dev corpora is only around 1000-2000 trees so we use it for validation
instead of including it in training).
As before, we convert the phrase structure trees to dependency trees using
Stanford Basic Dependencies. 
The parser and the mapper hyperparameters were tuned separately on the development set for
each domain. 
The unseen rate is typically 6-10\% in the domains of the Web Treebank.
We used the Stanford tagger~\cite{toutanova2003feature}, which was trained on the OntoNotes
training corpus, for part-of-speech tagging the Web Treebank corpora. The
tagger used bidirectional architecture and it
included word shape and distributional similarity features.
We train a separate mapper for each domain, tuning mapper hyperparameters
separately for each domain using the development sets. In this way, we obtain
some of the benefits of domain adaptation for each target domain.  

\paragraph{Downstream Task: Sentiment Analysis with Dependency Tree LSTMs}
We also perform experiments to analyze the effects of embedding mapping
on a downstream task, in this case sentiment analysis using the Stanford Sentiment Treebank~\cite{socher-EtAl:2013:EMNLP}. 
We use the dependency tree long short-term memory network (Tree-LSTM) proposed by
\newcite{tai2015improved}, simply replacing their default dependency parser with our version that maps unseen words. 
The dependency parser is trained on the WSJ corpus and mapped using the WSJ
development set. We use the same mapper that was optimized for  the WSJ
development set, without further hyperparameter tuning for the mapper. 
For the Tree-LSTM model, we use the same hyperparameter tuning as
described in \newcite{tai2015improved}. We use the standard train/development/test splits 
of 6820/872/1821 sentences for the binary classification task and 8544/1101/2210 for the
fine-grained task. 

\subsection{Mapper Settings and Hyperparameters}
The initial embeddings given to the mapper are the same as the initial embeddings given to
the parser. 
These are the 100-dimensional GloVe embeddings mentioned above. 
The output dimensionality of the mapper is also fixed to 100. 
All model parameters of the mappers are initialized to zero. 
We set the dimensionality of the non-linear layer to 400 across all experiments. 
The model parameters are optimized by maximizing the weighted multiple-loss objective
using L-BFGS with elastic-net regularization (Section~\ref{sec:theory}). The hyperparameters include the relative weight of the two objective terms
($\alpha$) and the regularization constants ($\lambda_1$, $\lambda_2$). For
$\alpha$, 
we search over values in $\{0, 0.1, 0.2, \ldots, 1\}$. 
For each of $\lambda_1$ and $\lambda_2$, we consider values in
$\{10^{-1}, 10^{-2}, \ldots, 10^{-9}, 0\}$. 
The hyperparameters are tuned via grid search to maximize the UAS on the development set.
\section{Results and Analysis}
\label{sec:results}

\begin{table*}[ht]
\centering
\begin{adjustbox}{max width=\textwidth}
\begin{tabular}{|l|c|c||c||c|c|c|c|c||c|}
\hline
\multicolumn{1}{|r}{}
&\multicolumn{3}{|c||}{Lower OOTV word rate} 
&\multicolumn{6}{c|}{Higher OOTV word rate}\\
\hline
& WSJ & OntoNotes & Avg. & Answers & Emails& Newsgroups & Reviews & Weblogs & Avg.\\
\hline
UAS  & 91.85$\rightarrow$92.21 & 90.17$\rightarrow$90.49 &90.38$\rightarrow$90.70 & 82.67$\rightarrow$83.21&81.76$\rightarrow$82.42&84.68$\rightarrow$85.13&84.25$\rightarrow$84.99&87.73$\rightarrow$88.43&84.22$\rightarrow$84.84\\
\hline
LAS& 89.49$\rightarrow$89.73 & 87.68$\rightarrow$87.92 &87.92$\rightarrow$88.14 &  78.98$\rightarrow$79.59 & 77.93$\rightarrow$78.56 & 81.88$\rightarrow$82.71 & 81.26$\rightarrow$81.92 & 85.68$\rightarrow$86.29 & 81.01$\rightarrow$81.81\\
\hline
OOTV \%& 2.72$\rightarrow$1.45 & 2.72$\rightarrow$1.4 &$-$  &8.53$\rightarrow$1.22 &10.56$\rightarrow$3.01 & 10.34$\rightarrow$1.04 & 6.84$\rightarrow$0.73 & 8.45$\rightarrow$0.38 & $-$\\
\hline
OOTV UAS& 89.88$\rightarrow$90.51 & 89.27$\rightarrow$89.81 &89.12$\rightarrow$89.78 & 80.88$\rightarrow$81.75 & 79.29$\rightarrow$81.02 &82.54$\rightarrow$83.71 & 81.17$\rightarrow$82.22 & 86.43$\rightarrow$87.31 &82.06$\rightarrow$83.20\\
\hline
\#Sents & 337 & 329 &$-$ & 671 & 644 & 579 & 632 & 541 & $-$\\
\hline
\end{tabular}
\end{adjustbox}
\caption{Results of dependency parsing 
on various treebanks.
Entries of the form A$\rightarrow$B give results for parsing without
mapped embeddings (A) and with mapped embeddings (B).  ``OOTV \%''
entries A$\rightarrow$B indicate that A\% of the test set vocabulary
was unseen in the parser training, and B\% remain unknown after
mapping the embeddings.  ``OOTV 
UAS'' refers to UAS measured on the subset of the test set sentences
that contain at least one OOTV word, 
and ``\#Sents'' gives the number of sentences in this subset.
}
\label{table:allres}
\end{table*}

\begin{figure*}
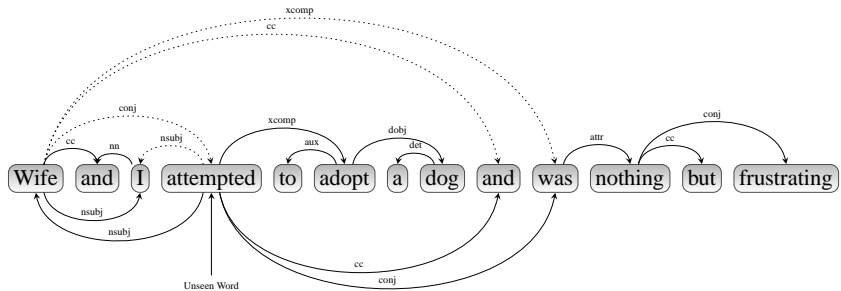
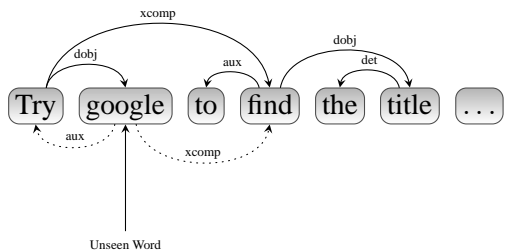
\centering
\begin{subfigure}{\textwidth}
  \centering
\adjustbox{max width=0.7\linewidth}{\begin{dependency}[theme = simple]
    \begin{deptext}[column sep=0.2cm, nodes={draw=gray!60!black, shade, top color=gray!60, rounded corners}]
    Wife \& and \& I \& attempted \& to \& adopt \& a \& dog \& and \& was \&
    nothing \& but \& frustrating\\
    \end{deptext}
    \depedge[edge below]{4}{1}{{nsubj}}
    \depedge[edge below]{1}{3}{{nsubj}}
    \depedge[edge below]{4}{9}{{cc}}
    \depedge[edge below]{4}{10}{{conj}}
    \depedge[edge style={dotted}]{4}{3}{{nsubj}}
    \depedge[edge style={dotted}]{1}{4}{{conj}}
    \depedge[edge style={dotted}]{1}{9}{{cc}}
    \depedge[edge style={dotted}]{1}{10}{{xcomp}}
    \depedge{8}{7}{{det}}
    \depedge{6}{8}{{dobj}}
    \depedge{1}{2}{{cc}}
    \depedge{6}{5}{{aux}}
    \depedge{4}{6}{{xcomp}}
    \depedge{3}{2}{{nn}}
    \depedge{11}{13}{{conj}}
    \depedge{10}{11}{{attr}}
    \deproot[edge below]{4}{{Unseen Word}}
    \depedge{11}{12}{{cc}}
\end{dependency}
}
\caption{We obtain correct attachments and correct tree after the mapper maps
the unseen word `attempted'.}
  \label{fig:ex1}
\end{subfigure}\\
\begin{subfigure}{\textwidth}
  \centering
\adjustbox{max width=\textwidth}{\begin{dependency}[theme = simple]
    \begin{deptext}[column sep=0.2cm, nodes={draw=gray!60!black, shade, top color=gray!60, rounded corners}]
Try \& google \& to \& find \& the \& title \& $\ldots$\\    \end{deptext}
    \depedge[edge below,edge style={dotted}]{2}{4}{{xcomp}}
    \depedge[edge below,edge style={dotted}]{2}{1}{{aux}}
    \depedge{1}{2}{{dobj}}
    \depedge{1}{4}{{xcomp}}
    \depedge{4}{3}{{aux}}
    \depedge{4}{6}{{dobj}}
    \depedge{6}{5}{{det}}
    \deproot[edge below]{2}{{Unseen Word}}
\end{dependency}
}
\caption{The mapper incorrectly maps `google', resulting in wrong
attachments and wrong tree.}
  \label{fig:ex2}
\end{subfigure}
\caption{Examples where the mapper helps and hurts: In the above examples the top 
arcs are 
before mapping and bottom 
ones are after mapping; dotted
lines refer to incorrect attachment.}
\label{fig:examples}
\end{figure*}

\subsection{Results on WSJ, OntoNotes, and Switchboard}
The upper half of Table~\ref{table:allres} shows our main test results on WSJ, OntoNotes, and Switchboard, the low-OOTV rate datasets.
Due to the small initial OOTV rates ($<$3\%), we only see modest gains of 0.3-0.4\% in UAS, 
with statistical significance at $p < 0.05$ for WSJ
and OntoNotes and $p < 0.07$ for Switchboard. 
The initial OOTV rates are cut in half by our mapper, with the remaining unknown words 
largely being numerical strings and misspellings.\footnote{We could potentially 
train the initial embeddings on a larger corpus or use heuristics to convert 
unknown numbers and misspellings to forms contained in our initial embeddings.} 
When only considering test sentences containing OOTV words 
(the row labeled ``OOTV subset''), the gains are significantly larger (0.5-0.8\% UAS at $p < 0.05$). 

\subsection{Results on Web Treebank}
The lower half of Table~\ref{table:allres} shows our main test results on the Web Treebank's five domains, the high-OOTV rate datasets.
As expected, the mapper has a much larger impact when parsing these 
out-of-domain datasets with high OOTV word rates.\footnote{As stated above, we train the parser on the OntoNotes dataset, but tune mapper hyperparameters  to maximize parsing performance on each development section of the Web Treebank's five domains. We then map the OOTV word vectors on each test set domain using the learned mapper for that domain.}

The OOTV rate reduction is much larger than for the WSJ-style datasets, and the parsing improvements (UAS and LAS) are statistically significant at  $p< 0.05$. 
On subsets containing at least one OOTV word (that also has an initial embedding), 
we see an average gain of $1.14\%$ UAS (see row labeled ``OOTV subset''). In this case, all improvements are statistically significant at $p < 0.02$. 
We observe that the relative reduction in OOTV\% for the Web Treebanks 
is larger than for the WSJ, OntoNotes, or Switchboard datasets. 
In particular, we are able to reduce the OOTV\% by 71-95\% relative. 
We also see the intuitive trend
that larger relative reductions in OOTV rate correlate with larger accuracy improvements. 
\subsection{Downstream Results} 
We now report results using the Dependency Tree-LSTM of \newcite{tai2015improved} 
for sentiment analysis on the Stanford Sentiment Treebank. 
We consider both the 
binary (positive/negative) and fine-grained classification tasks (\{very negative, negative, neutral,
positive, and very positive\}). 
We use the implementation provided by \newcite{tai2015improved}, changing only the 
dependency parses that are fed to their model. 
The sentiment dataset contains approximately 25\% OOTV words in the training set vocabulary,
5\% in the development set vocabulary, and 9\% in the test set vocabulary.
We map unseen words using the mapper tuned on the WSJ development set. 
We use the same Dependency Tree-LSTM experimental settings as Tai et al. \nocite{tai2015improved}
Results are shown in Table~\ref{table:lstm}. 
We improve upon the original accuracies in both binary and fine-grained
classification.~\footnote{Note that we report accuracies and improvements on the
dependency parse based system of~\newcite{tai2015improved} because the neural parser
that we use is dependency-based.} 
We also reduce the OOTV rate from 25\% in the training set vocabulary to about 6\%, and from 9\% in the 
test set vocabulary down to 4\%. 

\begin{table}[t]
\centering
\begin{adjustbox}{max width=\textwidth}
\begin{tabular}{|c|c|c|}
\hline
Fine-Grained  &  Binary\\
\hline
48.4$\rightarrow$49.5 &  85.7$\rightarrow$ 86.1 \\
\hline
\end{tabular}
\end{adjustbox}
\caption{Improvements on Stanford Sentiment Treebank test set using our parser with the Dependency Tree-LSTM.}
\label{table:lstm}
\end{table}

\subsection{Effect of Thresholds}

We also experimented with different values for the thresholds described in Section~\ref{sec:theory}. 
For the mapping threshold $\tau_m$, mapper-training threshold $\tau_t$, 
and parser threshold $\tau_p$, we consider the following four settings: 
\begin{align}
\boldsymbol{t_1}: & \:\tau_m = \tau_t = \tau_p = 1\nonumber\\
\boldsymbol{t_3}: & \:\tau_m = \tau_t = \tau_p = 3\nonumber\\
\boldsymbol{t_5}: & \:\tau_m = \tau_t = \tau_p = 5\nonumber\\
\boldsymbol{t_\infty}: & \:\tau_m = \infty, \tau_p = \tau_t = 5\nonumber
\end{align}
\noindent Using $\tau_m = \infty$ corresponds to mapping all words at test time, even 
words that we have seen many times in the training data and learned 
task-specific embeddings for.

We report the average development set UAS over all Web Treebank domains in Table~\ref{table:threshold}.
We see that $\boldsymbol{t_3}$ performs best, though settings $\boldsymbol{t_1}$ and 
$\boldsymbol{t_5}$ also improve over the baseline. At threshold $\boldsymbol{t_3}$ we have
approximately 20,000 examples for training the mapper, while at threshold $\boldsymbol{t_5}$ we have
only about 10,000 examples. We see a performance drop at $\boldsymbol{t_\infty}$, 
so it appears better to directly use the task-specific 
embeddings for words that appear frequently in the training data. In other results reported 
in this paper, we used $\boldsymbol{t_3}$ for the Web Treebank test sets and $\boldsymbol{t_1}$ for the rest. 

\begin{table}[t]
\centering
\begin{adjustbox}{max width=\textwidth}
\begin{tabular}{|c|c|c|c|c|}
\hline
Baseline & $\boldsymbol{t_1}$ & $\boldsymbol{t_3}$ & $\boldsymbol{t_5}$ & $\boldsymbol{t_\infty}$ \\
\hline
84.11 & 84.89 & 84.97 & 84.81 &84.14 \\\hline
\end{tabular}
\end{adjustbox}
\caption{Average Web Treebank development UAS at different threshold settings.} 
\label{table:threshold}
\end{table}

\subsection{Effect of Weighted Multi-Loss Objective}
We analyzed the results when varying $\alpha$, which balances between the two components 
of the mapper's multi-loss objective function. 
We found that, for all domains except Answers,
the best results are obtained with some $\alpha$ between 0 and 1. The optimal values outperformed the 
cases with $\alpha=0$ and $\alpha=1$ by 0.1-0.3\% UAS absolute. 
However, on the Answers domain, the best performance was achieved with $\alpha=0$; 
i.e., the mapper preferred mean squared error. For other domains, the optimal 
$\alpha$ tended to be within the range $[0.3,0.7]$.

\subsection{Comparison with Related
Work}
We compare to the approach presented by \newcite{tafforeau2015alr}. They propose
to refine embeddings for unseen words based on the relative shifts of their 
$k$ nearest
neighbors in the original embeddings space. 
Specifically, they define ``artificial refinement'' as: 
\begin{equation}
\phi_r(t) = \phi_o(t) + \sum_{k=1}^{K} \alpha_k (\phi_r(n_k) - \phi_o(n_k))
\end{equation}
where $\phi_r(.)$ is the vector in the refined embedding space and
$\phi_o(.)$ is the vector in the original embedding space. They define
$\alpha_k$ to be proportional to the cosine similarity between the target unseen
word ($t$) and neighbor ($n_k$): 
\[
\alpha_k = s(t,n_k) = \frac{\phi_o(t).\phi_o(n_k)}{|\phi(t)||\phi_o(n_k)|}
\]

\begin{table}[ht]
\centering
\begin{tabular}{|l|c|c|}
\hline
& Avg.~UAS & Avg.~LAS\\
\hline
Baseline &84.11 &81.02 \\
\hline
$k$-NN &84.54 &81.38 \\
\hline
Our Mapped & 84.97&81.79 \\ 
\hline
\end{tabular}
\caption{Comparison to $k$-nearest neighbor matching of Tafforeau et al. (2015).} \label{table:knn}
\end{table}
Table~\ref{table:knn} shows the average performance of the models over the development sets of the Web Treebank.  
On average, our mapper outperforms the $k$-NN approach ($k=3$).

\subsection{Dependency Parsing Examples}
In Figure~\ref{fig:examples}, we show two sentences: an instance where
the mapper helps and another where the mapper hurts the parsing
performance.\footnote{Sentences in Figure~\ref{fig:examples} are taken from 
the development portion of the Answers domain from the Web Treebank.}
In the first sentence (Figure~\ref{fig:ex1}), the parsing model has not seen the word
`attempted' during training. Note that the sentence contains 3 verbs: `attempted', `adopt',
and `was'. Even with the POS tags, the parser was unable to get the correct 
dependency attachment. After mapping, the parser correctly makes `attempted'
the root and gets the correct arcs and the correct tree. 
The 3 nearest neighbors of `attempted' in the mapped embedding space 
are `attempting', `tried', and `attempt'. We also see here that a single unseen 
word can lead to multiple errors in the parse. 

In the second example (Figure~\ref{fig:ex2}), the default model assigns the correct 
arcs using the POS  information even though it has not seen the word `google'.
However, using the mapped representation for `google', the parser makes errors. 
The 3-nearest neighbors for `google' in the mapped space are 
`damned', `look', and `hash'. We hypothesize that the mapper has
mapped this noun instance of `google' to be closer to verbs instead of nouns, which would explain the incorrect attachment. 

\subsection{Analyzing Mapped Representations}
\begin{figure}[h!]
\begin{subfigure}{.5\textwidth}
  \centering
  \includegraphics[width=\linewidth]{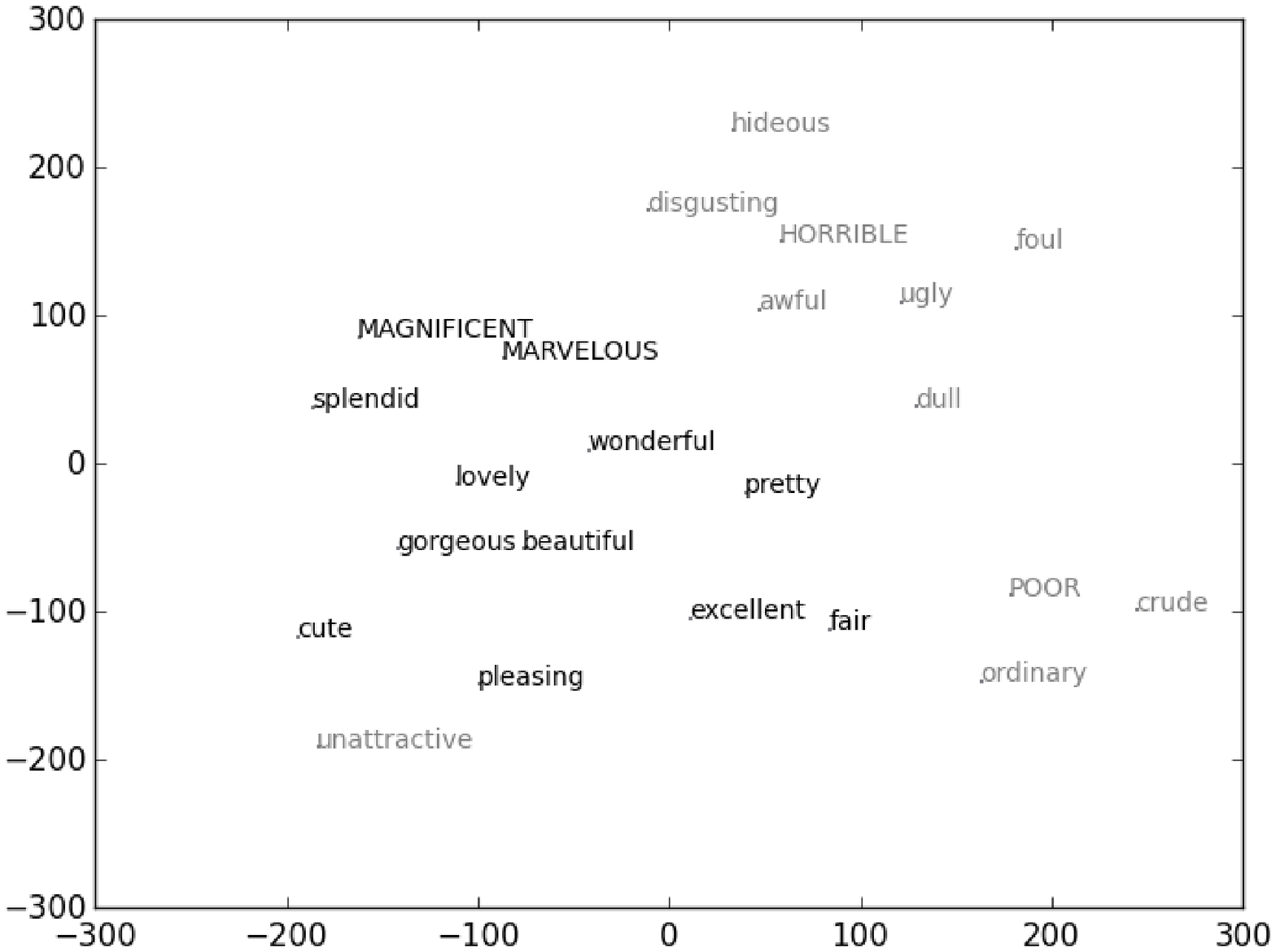}
  \caption{Initial Representational Space}
  \label{fig:sfig1}
\end{subfigure}\\
\begin{subfigure}{.5\textwidth}
  \centering
  \includegraphics[width=\linewidth]{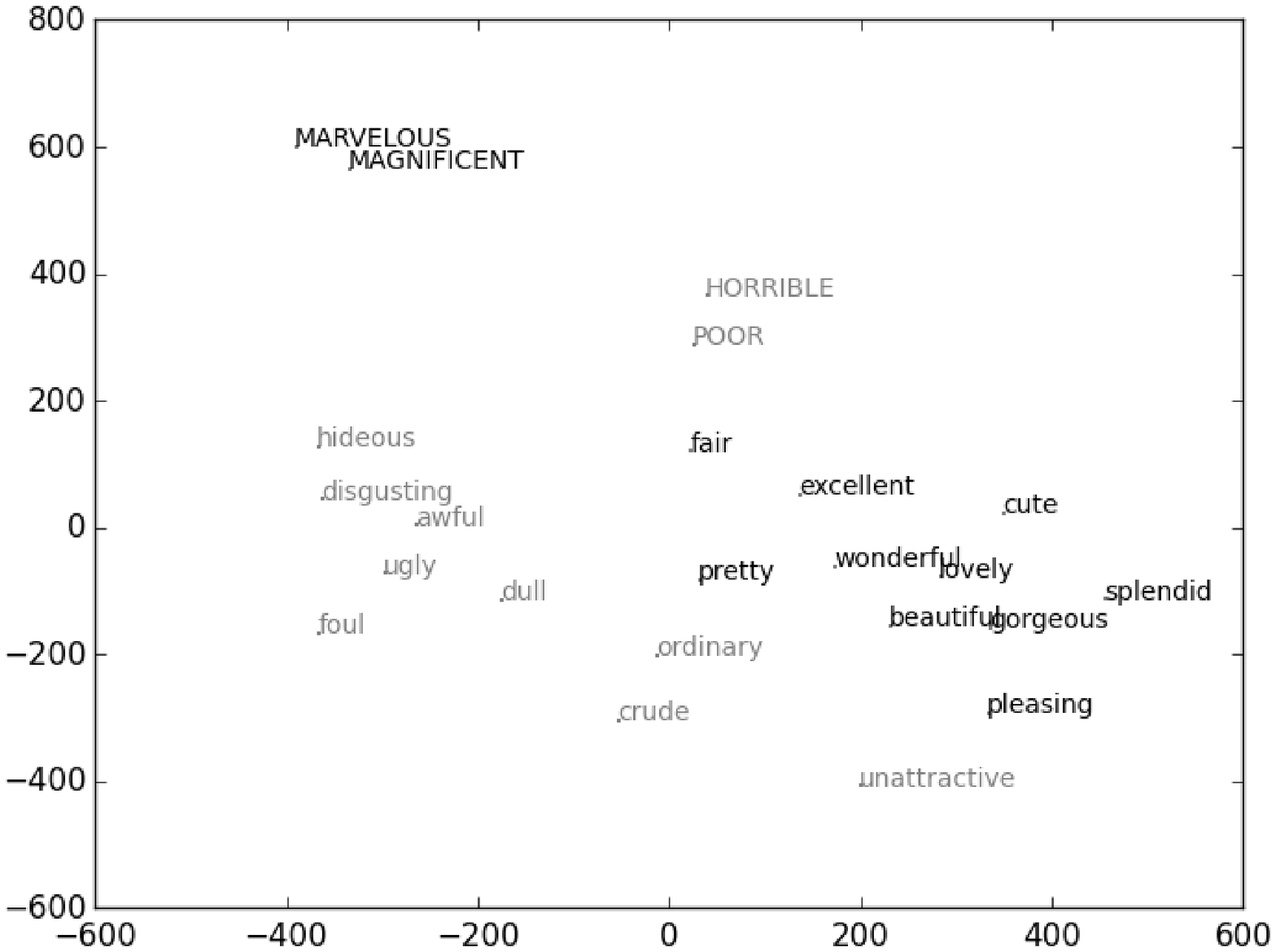}
  \caption{Learned Representational Space}
  \label{fig:sfig2}
\end{subfigure}\\
\begin{subfigure}{.5\textwidth}
  \centering
  \includegraphics[width=\linewidth]{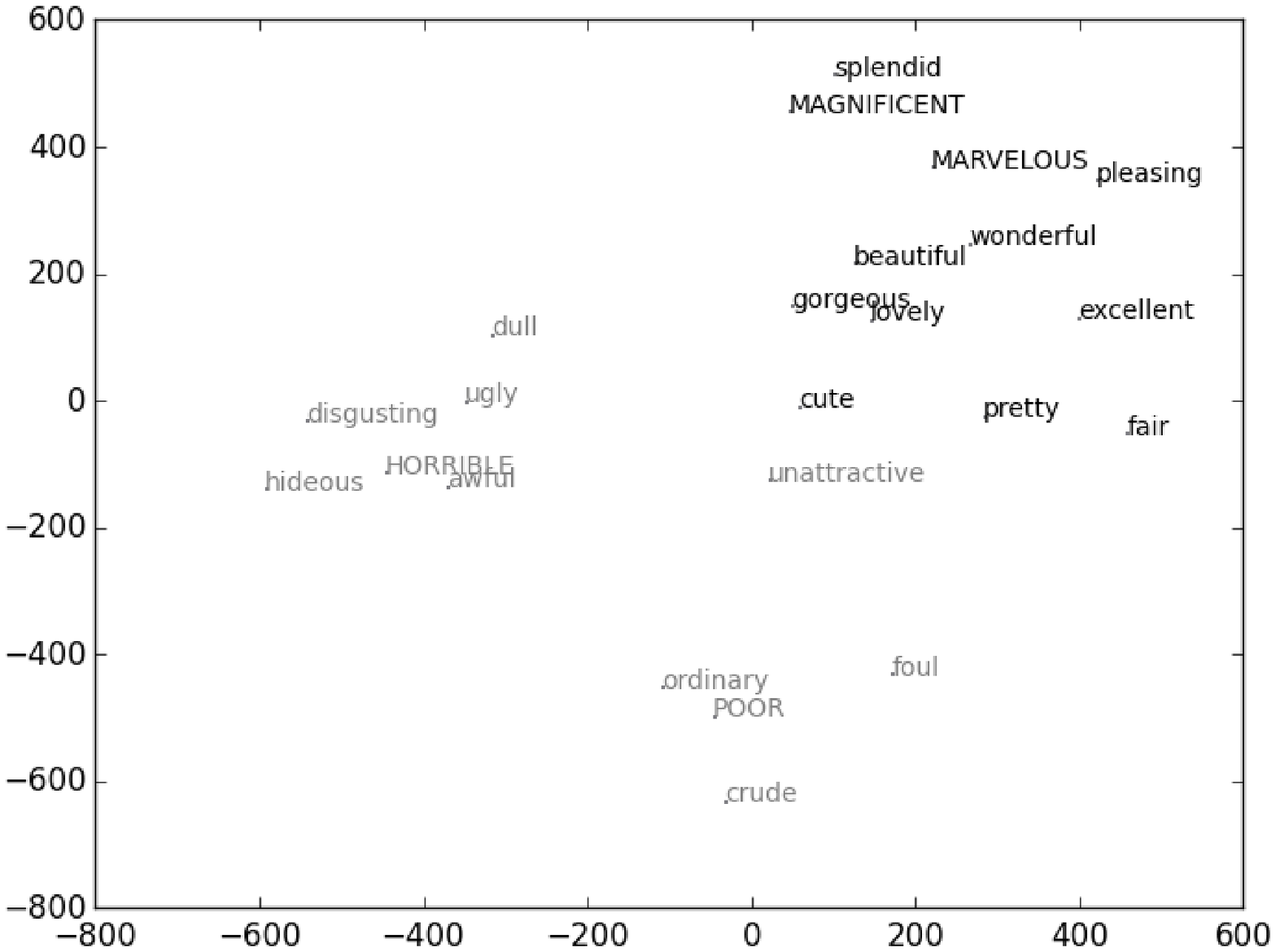}
  \caption{Mapped Representational Space}
  \label{fig:sfig3}
\end{subfigure}
\caption{t-SNE plots on initial, parser trained, and mapped
representations. 
}\label{fig:tsne}
\end{figure}

To understand the mapped embedding space,
we use t-SNE~\cite{van2008visualizing} to visualize a small subset of embeddings. 
In Figure~\ref{fig:tsne}, we plot the initial embeddings, the
parser-trained embeddings, and finally the mapped embeddings. We include
four unseen words (shown in caps): `horrible', `poor', `marvelous', and `magnificent'. 
In Figure~\ref{fig:sfig1} and Figure~\ref{fig:sfig2}, the embeddings for 
the unseen words are identical (even though t-SNE places them in different places when producing its projection). 
In Figure~\ref{fig:sfig3}, we observe that the mapper has placed the unseen words 
within appropriate areas of the space with respect to similarity with the seen words. 
We contrast this with Figure~\ref{fig:sfig2}, in which the unseen words appear to be 
within a different region of the space from all seen words. 

\section{Conclusion}
We have described a simple method to resolve unseen
words when training supervised models that learn task-specific word embeddings:  
a feed-forward neural network that maps initial embeddings to the task-specific 
embedding space. 
We demonstrated significant improvements in dependency parsing accuracy across
several domains, as well as improvements on a downstream task. 
Our approach is simple, effective, and applicable to many other settings, both
inside and outside NLP. 

\section*{Acknowledgments}
We would like to thank the anonymous reviewers for their useful comments. This
research was supported by a Google Faculty Research Award to Mohit Bansal,
Karen Livescu and Kevin Gimpel.

\bibliographystyle{acl2016}
\bibliography{refs}

\end{document}

%% file: figs/learning.tex
\ifx\setlinejoinmode\undefined
  \newcommand{\setlinejoinmode}[1]{}
\fi
\ifx\setlinecaps\undefined
  \newcommand{\setlinecaps}[1]{}
\fi
\ifx\setfont\undefined
  \newcommand{\setfont}[2]{}
\fi
\pspicture(34.908825,-17.960018)(58.005000,-2.029210)
\psscalebox{1.000000 -1.000000}{
\newrgbcolor{dialinecolor}{0.000000 0.000000 0.000000}%
\psset{linecolor=dialinecolor}
\newrgbcolor{diafillcolor}{1.000000 1.000000 1.000000}%
\psset{fillcolor=diafillcolor}
\psset{linewidth=0.030000cm}
\psset{linestyle=solid}
\psset{linestyle=solid}
\setlinejoinmode{0}
\setlinecaps{0}
\newrgbcolor{dialinecolor}{0.000000 0.000000 0.000000}%
\psset{linecolor=dialinecolor}
\psline(36.188825,6.287500)(36.183333,6.287500)(36.183333,13.912500)(38.730284,13.912500)
\psset{linestyle=solid}
\setlinejoinmode{0}
\setlinecaps{0}
\newrgbcolor{dialinecolor}{0.000000 0.000000 0.000000}%
\psset{linecolor=dialinecolor}
\pspolygon*(38.880284,13.912500)(38.680284,14.012500)(38.730284,13.912500)(38.680284,13.812500)
\newrgbcolor{dialinecolor}{0.000000 0.000000 0.000000}%
\psset{linecolor=dialinecolor}
\pspolygon(38.880284,13.912500)(38.680284,14.012500)(38.730284,13.912500)(38.680284,13.812500)
\psset{linewidth=0.030000cm}
\psset{linestyle=solid}
\psset{linestyle=solid}
\setlinejoinmode{0}
\setlinecaps{0}
\newrgbcolor{dialinecolor}{0.000000 0.000000 0.000000}%
\psset{linecolor=dialinecolor}
\psline(46.051225,6.175000)(46.051225,7.904167)(49.516667,7.904167)(49.516667,9.449792)
\psset{linestyle=solid}
\setlinejoinmode{0}
\setlinecaps{0}
\newrgbcolor{dialinecolor}{0.000000 0.000000 0.000000}%
\psset{linecolor=dialinecolor}
\pspolygon*(49.516667,9.599792)(49.416667,9.399792)(49.516667,9.449792)(49.616667,9.399792)
\newrgbcolor{dialinecolor}{0.000000 0.000000 0.000000}%
\psset{linecolor=dialinecolor}
\pspolygon(49.516667,9.599792)(49.416667,9.399792)(49.516667,9.449792)(49.616667,9.399792)
\setfont{Helvetica}{1.058333}
\newrgbcolor{dialinecolor}{0.000000 0.000000 0.000000}%
\psset{linecolor=dialinecolor}
\rput[l](48.950000,10.333333){\psscalebox{1 -1}{$loss(e_{i}^{t}, e_{i}^{m})$}}
\setfont{Helvetica}{0.800000}
\newrgbcolor{dialinecolor}{0.000000 0.000000 0.000000}%
\psset{linecolor=dialinecolor}
\rput[l](47.820075,5.202221){\psscalebox{1 -1}{}}
\psset{linewidth=0.030000cm}
\psset{linestyle=solid}
\psset{linestyle=solid}
\setlinecaps{0}
\newrgbcolor{dialinecolor}{0.000000 0.000000 0.000000}%
\psset{linecolor=dialinecolor}
\psline(38.676581,3.188575)(39.793033,3.169325)
\psset{linestyle=solid}
\setlinejoinmode{0}
\setlinecaps{0}
\newrgbcolor{dialinecolor}{0.000000 0.000000 0.000000}%
\psset{linecolor=dialinecolor}
\pspolygon*(39.950510,3.166610)(39.742351,3.275215)(39.793033,3.169325)(39.738731,3.065246)
\newrgbcolor{dialinecolor}{0.000000 0.000000 0.000000}%
\psset{linecolor=dialinecolor}
\pspolygon(39.950510,3.166610)(39.742351,3.275215)(39.793033,3.169325)(39.738731,3.065246)
\psset{linewidth=0.030000cm}
\psset{linestyle=solid}
\psset{linestyle=solid}
\setlinecaps{0}
\setlinejoinmode{0}
\psset{linewidth=0.030000cm}
\setlinecaps{0}
\setlinejoinmode{0}
\psset{linestyle=solid}
\newrgbcolor{dialinecolor}{1.000000 1.000000 1.000000}%
\psset{linecolor=dialinecolor}
\pscustom{
\newpath
\moveto(45.060700,2.044210)
\lineto(48.120075,2.044210)
\curveto(47.661169,2.367710)(47.508200,2.529460)(47.508200,2.852960)
\curveto(47.508200,3.176460)(47.661169,3.338210)(48.120075,3.661710)
\lineto(45.060700,3.661710)
\curveto(44.601794,3.338210)(44.448825,3.176460)(44.448825,2.852960)
\curveto(44.448825,2.529460)(44.601794,2.367710)(45.060700,2.044210)
\fill[fillstyle=solid,fillcolor=diafillcolor,linecolor=diafillcolor]}
\newrgbcolor{dialinecolor}{0.000000 0.000000 0.000000}%
\psset{linecolor=dialinecolor}
\pscustom{
\newpath
\moveto(45.060700,2.044210)
\lineto(48.120075,2.044210)
\curveto(47.661169,2.367710)(47.508200,2.529460)(47.508200,2.852960)
\curveto(47.508200,3.176460)(47.661169,3.338210)(48.120075,3.661710)
\lineto(45.060700,3.661710)
\curveto(44.601794,3.338210)(44.448825,3.176460)(44.448825,2.852960)
\curveto(44.448825,2.529460)(44.601794,2.367710)(45.060700,2.044210)
\stroke}
\setfont{Helvetica}{0.529167}
\newrgbcolor{dialinecolor}{0.000000 0.000000 0.000000}%
\psset{linecolor=dialinecolor}
\rput(46.284450,2.456085){\psscalebox{1 -1}{Model}}
\setfont{Helvetica}{0.529167}
\newrgbcolor{dialinecolor}{0.000000 0.000000 0.000000}%
\psset{linecolor=dialinecolor}
\rput(46.284450,2.985251){\psscalebox{1 -1}{Parameters}}
\setfont{Helvetica}{0.529167}
\newrgbcolor{dialinecolor}{0.000000 0.000000 0.000000}%
\psset{linecolor=dialinecolor}
\rput(46.284450,3.514418){\psscalebox{1 -1}{}}
\psset{linewidth=0.030000cm}
\psset{linestyle=solid}
\psset{linestyle=solid}
\setlinecaps{0}
\newrgbcolor{dialinecolor}{0.000000 0.000000 0.000000}%
\psset{linecolor=dialinecolor}
\psline(38.714081,5.463575)(39.830533,5.444325)
\psset{linestyle=solid}
\setlinejoinmode{0}
\setlinecaps{0}
\newrgbcolor{dialinecolor}{0.000000 0.000000 0.000000}%
\psset{linecolor=dialinecolor}
\pspolygon*(39.988010,5.441610)(39.779851,5.550215)(39.830533,5.444325)(39.776231,5.340246)
\newrgbcolor{dialinecolor}{0.000000 0.000000 0.000000}%
\psset{linecolor=dialinecolor}
\pspolygon(39.988010,5.441610)(39.779851,5.550215)(39.830533,5.444325)(39.776231,5.340246)
\setfont{Helvetica}{0.705556}
\newrgbcolor{dialinecolor}{0.000000 0.000000 0.000000}%
\psset{linecolor=dialinecolor}
\rput[l](45.936325,3.431710){\psscalebox{1 -1}{$W$}}
\newrgbcolor{dialinecolor}{1.000000 1.000000 1.000000}%
\psset{linecolor=dialinecolor}
\pspolygon*(35.976191,2.172500)(38.918994,2.172500)(38.330272,3.790000)(35.387469,3.790000)
\psset{linewidth=0.030000cm}
\psset{linestyle=solid}
\psset{linestyle=solid}
\setlinejoinmode{0}
\newrgbcolor{dialinecolor}{0.000000 0.000000 0.000000}%
\psset{linecolor=dialinecolor}
\pspolygon(35.976191,2.172500)(38.918994,2.172500)(38.330272,3.790000)(35.387469,3.790000)
\setfont{Helvetica}{0.529167}
\newrgbcolor{dialinecolor}{0.000000 0.000000 0.000000}%
\psset{linecolor=dialinecolor}
\rput(37.153232,2.580000){\psscalebox{1 -1}{Annotated }}
\setfont{Helvetica}{0.529167}
\newrgbcolor{dialinecolor}{0.000000 0.000000 0.000000}%
\psset{linecolor=dialinecolor}
\rput(37.153232,3.109167){\psscalebox{1 -1}{Training}}
\setfont{Helvetica}{0.529167}
\newrgbcolor{dialinecolor}{0.000000 0.000000 0.000000}%
\psset{linecolor=dialinecolor}
\rput(37.153232,3.438333){\psscalebox{1 -1}{Sentences}}
\newrgbcolor{dialinecolor}{1.000000 1.000000 1.000000}%
\psset{linecolor=dialinecolor}
\pspolygon*(35.601720,4.350000)(38.824522,4.350000)(38.146628,6.212500)(34.923825,6.212500)
\psset{linewidth=0.030000cm}
\psset{linestyle=solid}
\psset{linestyle=solid}
\setlinejoinmode{0}
\newrgbcolor{dialinecolor}{0.000000 0.000000 0.000000}%
\psset{linecolor=dialinecolor}
\pspolygon(35.601720,4.350000)(38.824522,4.350000)(38.146628,6.212500)(34.923825,6.212500)
\setfont{Helvetica}{0.529167}
\newrgbcolor{dialinecolor}{0.000000 0.000000 0.000000}%
\psset{linecolor=dialinecolor}
\rput(36.874174,4.880000){\psscalebox{1 -1}{Initial}}
\setfont{Helvetica}{0.529167}
\newrgbcolor{dialinecolor}{0.000000 0.000000 0.000000}%
\psset{linecolor=dialinecolor}
\rput(36.874174,5.409167){\psscalebox{1 -1}{Embeddings}}
\setfont{Helvetica}{0.529167}
\newrgbcolor{dialinecolor}{0.000000 0.000000 0.000000}%
\psset{linecolor=dialinecolor}
\rput(36.874174,5.938333){\psscalebox{1 -1}{}}
\setfont{Helvetica}{0.776111}
\newrgbcolor{dialinecolor}{0.000000 0.000000 0.000000}%
\psset{linecolor=dialinecolor}
\rput[l](36.321325,5.922500){\psscalebox{1 -1}{$e_{i}^{o}$}}
\psset{linewidth=0.030000cm}
\psset{linestyle=solid}
\psset{linestyle=solid}
\setlinecaps{0}
\newrgbcolor{dialinecolor}{0.000000 0.000000 0.000000}%
\psset{linecolor=dialinecolor}
\psline(43.398981,3.004365)(44.185186,3.000853)
\psset{linestyle=solid}
\setlinejoinmode{0}
\setlinecaps{0}
\newrgbcolor{dialinecolor}{0.000000 0.000000 0.000000}%
\psset{linecolor=dialinecolor}
\pspolygon*(44.342684,3.000150)(44.133155,3.106087)(44.185186,3.000853)(44.132217,2.896089)
\newrgbcolor{dialinecolor}{0.000000 0.000000 0.000000}%
\psset{linecolor=dialinecolor}
\pspolygon(44.342684,3.000150)(44.133155,3.106087)(44.185186,3.000853)(44.132217,2.896089)
\psset{linewidth=0.030000cm}
\psset{linestyle=solid}
\psset{linestyle=solid}
\setlinecaps{0}
\newrgbcolor{dialinecolor}{0.000000 0.000000 0.000000}%
\psset{linecolor=dialinecolor}
\psline(43.573792,4.926817)(44.460184,4.925322)
\psset{linestyle=solid}
\setlinejoinmode{0}
\setlinecaps{0}
\newrgbcolor{dialinecolor}{0.000000 0.000000 0.000000}%
\psset{linecolor=dialinecolor}
\pspolygon*(44.617684,4.925057)(44.407861,5.030411)(44.460184,4.925322)(44.407507,4.820411)
\newrgbcolor{dialinecolor}{0.000000 0.000000 0.000000}%
\psset{linecolor=dialinecolor}
\pspolygon(44.617684,4.925057)(44.407861,5.030411)(44.460184,4.925322)(44.407507,4.820411)
\newrgbcolor{dialinecolor}{1.000000 1.000000 1.000000}%
\psset{linecolor=dialinecolor}
\pspolygon*(39.112778,11.693132)(39.112778,16.729798)(39.628534,16.729798)(39.628534,11.693132)
\psset{linewidth=0.030000cm}
\psset{linestyle=solid}
\psset{linestyle=solid}
\setlinejoinmode{0}
\newrgbcolor{dialinecolor}{0.000000 0.000000 0.000000}%
\psset{linecolor=dialinecolor}
\pspolygon(39.112778,11.693132)(39.112778,16.729798)(39.628534,16.729798)(39.628534,11.693132)
\setfont{Helvetica}{0.800000}
\newrgbcolor{dialinecolor}{0.000000 0.000000 0.000000}%
\psset{linecolor=dialinecolor}
\rput(39.370656,14.406465){\psscalebox{1 -1}{}}
\psset{linewidth=0.030000cm}
\psset{linestyle=solid}
\psset{linestyle=solid}
\setlinecaps{0}
\newrgbcolor{dialinecolor}{0.000000 0.000000 0.000000}%
\psset{linecolor=dialinecolor}
\psline(39.662778,12.865000)(40.266760,12.874587)
\psset{linestyle=solid}
\setlinejoinmode{0}
\setlinecaps{0}
\newrgbcolor{dialinecolor}{0.000000 0.000000 0.000000}%
\psset{linecolor=dialinecolor}
\pspolygon*(40.416741,12.876968)(40.215179,12.973781)(40.266760,12.874587)(40.218353,12.773806)
\newrgbcolor{dialinecolor}{0.000000 0.000000 0.000000}%
\psset{linecolor=dialinecolor}
\pspolygon(40.416741,12.876968)(40.215179,12.973781)(40.266760,12.874587)(40.218353,12.773806)
\newrgbcolor{dialinecolor}{1.000000 1.000000 1.000000}%
\psset{linecolor=dialinecolor}
\pspolygon*(43.922778,11.690000)(43.922778,16.726667)(44.438534,16.726667)(44.438534,11.690000)
\psset{linewidth=0.030000cm}
\psset{linestyle=solid}
\psset{linestyle=solid}
\setlinejoinmode{0}
\newrgbcolor{dialinecolor}{0.000000 0.000000 0.000000}%
\psset{linecolor=dialinecolor}
\pspolygon(43.922778,11.690000)(43.922778,16.726667)(44.438534,16.726667)(44.438534,11.690000)
\setfont{Helvetica}{0.800000}
\newrgbcolor{dialinecolor}{0.000000 0.000000 0.000000}%
\psset{linecolor=dialinecolor}
\rput(44.180656,14.403333){\psscalebox{1 -1}{}}
\psset{linewidth=0.030000cm}
\psset{linestyle=solid}
\psset{linestyle=solid}
\setlinecaps{0}
\newrgbcolor{dialinecolor}{0.000000 0.000000 0.000000}%
\psset{linecolor=dialinecolor}
\psline(39.673014,13.762231)(40.276996,13.771818)
\psset{linestyle=solid}
\setlinejoinmode{0}
\setlinecaps{0}
\newrgbcolor{dialinecolor}{0.000000 0.000000 0.000000}%
\psset{linecolor=dialinecolor}
\pspolygon*(40.426977,13.774198)(40.225415,13.871012)(40.276996,13.771818)(40.228589,13.671037)
\newrgbcolor{dialinecolor}{0.000000 0.000000 0.000000}%
\psset{linecolor=dialinecolor}
\pspolygon(40.426977,13.774198)(40.225415,13.871012)(40.276996,13.771818)(40.228589,13.671037)
\psset{linewidth=0.030000cm}
\psset{linestyle=solid}
\psset{linestyle=solid}
\setlinecaps{0}
\newrgbcolor{dialinecolor}{0.000000 0.000000 0.000000}%
\psset{linecolor=dialinecolor}
\psline(39.648014,14.537231)(40.251996,14.546818)
\psset{linestyle=solid}
\setlinejoinmode{0}
\setlinecaps{0}
\newrgbcolor{dialinecolor}{0.000000 0.000000 0.000000}%
\psset{linecolor=dialinecolor}
\pspolygon*(40.401977,14.549198)(40.200415,14.646012)(40.251996,14.546818)(40.203589,14.446037)
\newrgbcolor{dialinecolor}{0.000000 0.000000 0.000000}%
\psset{linecolor=dialinecolor}
\pspolygon(40.401977,14.549198)(40.200415,14.646012)(40.251996,14.546818)(40.203589,14.446037)
\psset{linewidth=0.030000cm}
\psset{linestyle=solid}
\psset{linestyle=solid}
\setlinecaps{0}
\newrgbcolor{dialinecolor}{0.000000 0.000000 0.000000}%
\psset{linecolor=dialinecolor}
\psline(39.668014,15.347231)(40.271996,15.356818)
\psset{linestyle=solid}
\setlinejoinmode{0}
\setlinecaps{0}
\newrgbcolor{dialinecolor}{0.000000 0.000000 0.000000}%
\psset{linecolor=dialinecolor}
\pspolygon*(40.421977,15.359198)(40.220415,15.456012)(40.271996,15.356818)(40.223589,15.256037)
\newrgbcolor{dialinecolor}{0.000000 0.000000 0.000000}%
\psset{linecolor=dialinecolor}
\pspolygon(40.421977,15.359198)(40.220415,15.456012)(40.271996,15.356818)(40.223589,15.256037)
\psset{linewidth=0.030000cm}
\psset{linestyle=solid}
\psset{linestyle=solid}
\setlinecaps{0}
\newrgbcolor{dialinecolor}{0.000000 0.000000 0.000000}%
\psset{linecolor=dialinecolor}
\psline(39.673014,16.162231)(40.276996,16.171818)
\psset{linestyle=solid}
\setlinejoinmode{0}
\setlinecaps{0}
\newrgbcolor{dialinecolor}{0.000000 0.000000 0.000000}%
\psset{linecolor=dialinecolor}
\pspolygon*(40.426977,16.174198)(40.225415,16.271012)(40.276996,16.171818)(40.228589,16.071037)
\newrgbcolor{dialinecolor}{0.000000 0.000000 0.000000}%
\psset{linecolor=dialinecolor}
\pspolygon(40.426977,16.174198)(40.225415,16.271012)(40.276996,16.171818)(40.228589,16.071037)
\psset{linewidth=0.030000cm}
\psset{linestyle=solid}
\psset{linestyle=solid}
\setlinecaps{0}
\newrgbcolor{dialinecolor}{0.000000 0.000000 0.000000}%
\psset{linecolor=dialinecolor}
\psline(43.060514,12.849731)(43.664496,12.859318)
\psset{linestyle=solid}
\setlinejoinmode{0}
\setlinecaps{0}
\newrgbcolor{dialinecolor}{0.000000 0.000000 0.000000}%
\psset{linecolor=dialinecolor}
\pspolygon*(43.814477,12.861698)(43.612915,12.958512)(43.664496,12.859318)(43.616089,12.758537)
\newrgbcolor{dialinecolor}{0.000000 0.000000 0.000000}%
\psset{linecolor=dialinecolor}
\pspolygon(43.814477,12.861698)(43.612915,12.958512)(43.664496,12.859318)(43.616089,12.758537)
\psset{linewidth=0.030000cm}
\psset{linestyle=solid}
\psset{linestyle=solid}
\setlinecaps{0}
\newrgbcolor{dialinecolor}{0.000000 0.000000 0.000000}%
\psset{linecolor=dialinecolor}
\psline(43.018014,13.734731)(43.621996,13.744318)
\psset{linestyle=solid}
\setlinejoinmode{0}
\setlinecaps{0}
\newrgbcolor{dialinecolor}{0.000000 0.000000 0.000000}%
\psset{linecolor=dialinecolor}
\pspolygon*(43.771977,13.746698)(43.570415,13.843512)(43.621996,13.744318)(43.573589,13.643537)
\newrgbcolor{dialinecolor}{0.000000 0.000000 0.000000}%
\psset{linecolor=dialinecolor}
\pspolygon(43.771977,13.746698)(43.570415,13.843512)(43.621996,13.744318)(43.573589,13.643537)
\psset{linewidth=0.030000cm}
\psset{linestyle=solid}
\psset{linestyle=solid}
\setlinecaps{0}
\newrgbcolor{dialinecolor}{0.000000 0.000000 0.000000}%
\psset{linecolor=dialinecolor}
\psline(43.025514,14.507231)(43.629496,14.516818)
\psset{linestyle=solid}
\setlinejoinmode{0}
\setlinecaps{0}
\newrgbcolor{dialinecolor}{0.000000 0.000000 0.000000}%
\psset{linecolor=dialinecolor}
\pspolygon*(43.779477,14.519198)(43.577915,14.616012)(43.629496,14.516818)(43.581089,14.416037)
\newrgbcolor{dialinecolor}{0.000000 0.000000 0.000000}%
\psset{linecolor=dialinecolor}
\pspolygon(43.779477,14.519198)(43.577915,14.616012)(43.629496,14.516818)(43.581089,14.416037)
\psset{linewidth=0.030000cm}
\psset{linestyle=solid}
\psset{linestyle=solid}
\setlinecaps{0}
\newrgbcolor{dialinecolor}{0.000000 0.000000 0.000000}%
\psset{linecolor=dialinecolor}
\psline(43.033014,15.304731)(43.636996,15.314318)
\psset{linestyle=solid}
\setlinejoinmode{0}
\setlinecaps{0}
\newrgbcolor{dialinecolor}{0.000000 0.000000 0.000000}%
\psset{linecolor=dialinecolor}
\pspolygon*(43.786977,15.316698)(43.585415,15.413512)(43.636996,15.314318)(43.588589,15.213537)
\newrgbcolor{dialinecolor}{0.000000 0.000000 0.000000}%
\psset{linecolor=dialinecolor}
\pspolygon(43.786977,15.316698)(43.585415,15.413512)(43.636996,15.314318)(43.588589,15.213537)
\psset{linewidth=0.030000cm}
\psset{linestyle=solid}
\psset{linestyle=solid}
\setlinecaps{0}
\newrgbcolor{dialinecolor}{0.000000 0.000000 0.000000}%
\psset{linecolor=dialinecolor}
\psline(43.053014,16.152231)(43.656996,16.161818)
\psset{linestyle=solid}
\setlinejoinmode{0}
\setlinecaps{0}
\newrgbcolor{dialinecolor}{0.000000 0.000000 0.000000}%
\psset{linecolor=dialinecolor}
\pspolygon*(43.806977,16.164198)(43.605415,16.261012)(43.656996,16.161818)(43.608589,16.061037)
\newrgbcolor{dialinecolor}{0.000000 0.000000 0.000000}%
\psset{linecolor=dialinecolor}
\pspolygon(43.806977,16.164198)(43.605415,16.261012)(43.656996,16.161818)(43.608589,16.061037)
\newrgbcolor{dialinecolor}{1.000000 1.000000 1.000000}%
\psset{linecolor=dialinecolor}
\pspolygon*(40.535278,10.715000)(40.535278,17.515000)(43.000694,17.515000)(43.000694,10.715000)
\psset{linewidth=0.030000cm}
\psset{linestyle=solid}
\psset{linestyle=solid}
\setlinejoinmode{0}
\newrgbcolor{dialinecolor}{0.000000 0.000000 0.000000}%
\psset{linecolor=dialinecolor}
\pspolygon(40.535278,10.715000)(40.535278,17.515000)(43.000694,17.515000)(43.000694,10.715000)
\setfont{Helvetica}{0.531989}
\newrgbcolor{dialinecolor}{0.000000 0.000000 0.000000}%
\psset{linecolor=dialinecolor}
\rput(41.767986,13.978011){\psscalebox{1 -1}{Non-Linear}}
\setfont{Helvetica}{0.531989}
\newrgbcolor{dialinecolor}{0.000000 0.000000 0.000000}%
\psset{linecolor=dialinecolor}
\rput(41.767986,14.510000){\psscalebox{1 -1}{Layer}}
\psset{linewidth=0.010000cm}
\psset{linestyle=dashed,dash=1.000000 1.000000}
\psset{linestyle=dashed,dash=1.000000 1.000000}
\setlinecaps{0}
\newrgbcolor{dialinecolor}{0.000000 0.000000 0.000000}%
\psset{linecolor=dialinecolor}
\psline(38.487778,9.565000)(38.475278,17.902500)
\psset{linewidth=0.010000cm}
\psset{linestyle=dashed,dash=1.000000 1.000000}
\psset{linestyle=dashed,dash=1.000000 1.000000}
\setlinecaps{0}
\newrgbcolor{dialinecolor}{0.000000 0.000000 0.000000}%
\psset{linecolor=dialinecolor}
\psline(45.525278,9.515000)(45.525278,17.902500)
\psset{linewidth=0.010000cm}
\psset{linestyle=dashed,dash=1.000000 1.000000}
\psset{linestyle=dashed,dash=1.000000 1.000000}
\setlinecaps{0}
\newrgbcolor{dialinecolor}{0.000000 0.000000 0.000000}%
\psset{linecolor=dialinecolor}
\psline(38.500278,9.552500)(45.512778,9.540000)
\psset{linewidth=0.010000cm}
\psset{linestyle=dashed,dash=1.000000 1.000000}
\psset{linestyle=dashed,dash=1.000000 1.000000}
\setlinecaps{0}
\newrgbcolor{dialinecolor}{0.000000 0.000000 0.000000}%
\psset{linecolor=dialinecolor}
\psline(38.487787,17.955009)(45.500287,17.942509)
\setfont{Helvetica}{0.705556}
\newrgbcolor{dialinecolor}{0.000000 0.000000 0.000000}%
\psset{linecolor=dialinecolor}
\rput[l](40.337778,10.277500){\psscalebox{1 -1}{Mapper Function}}
\psset{linewidth=0.030000cm}
\psset{linestyle=solid}
\psset{linestyle=solid}
\setlinejoinmode{0}
\setlinecaps{0}
\newrgbcolor{dialinecolor}{0.000000 0.000000 0.000000}%
\psset{linecolor=dialinecolor}
\pscustom{
\newpath
\moveto(40.787778,13.302500)
\curveto(41.713228,13.302500)(41.862328,11.877500)(42.787778,11.877500)
\stroke}
\psset{linewidth=0.000100cm}
\psset{linestyle=solid}
\psset{linestyle=solid}
\setlinecaps{0}
\newrgbcolor{dialinecolor}{0.000000 0.000000 0.000000}%
\psset{linecolor=dialinecolor}
\psline(40.963001,12.627500)(42.575054,12.627500)
\psset{linewidth=0.000100cm}
\psset{linestyle=solid}
\setlinejoinmode{0}
\setlinecaps{0}
\newrgbcolor{dialinecolor}{0.000000 0.000000 0.000000}%
\psset{linecolor=dialinecolor}
\psline(41.072890,12.572500)(40.962890,12.627500)(41.072890,12.682500)
\psset{linewidth=0.000100cm}
\psset{linestyle=solid}
\setlinejoinmode{0}
\setlinecaps{0}
\newrgbcolor{dialinecolor}{0.000000 0.000000 0.000000}%
\psset{linecolor=dialinecolor}
\psline(42.465166,12.682500)(42.575166,12.627500)(42.465166,12.572500)
\psset{linewidth=0.000000cm}
\psset{linestyle=solid}
\psset{linestyle=solid}
\setlinecaps{0}
\newrgbcolor{dialinecolor}{0.000000 0.000000 0.000000}%
\psset{linecolor=dialinecolor}
\psline(41.745278,11.450000)(41.750278,13.602500)
\psset{linewidth=0.000000cm}
\psset{linestyle=solid}
\setlinejoinmode{0}
\setlinecaps{0}
\newrgbcolor{dialinecolor}{0.000000 0.000000 0.000000}%
\psset{linecolor=dialinecolor}
\psline(41.800533,11.559872)(41.745278,11.450000)(41.690533,11.560127)
\psset{linewidth=0.000000cm}
\psset{linestyle=solid}
\setlinejoinmode{0}
\setlinecaps{0}
\newrgbcolor{dialinecolor}{0.000000 0.000000 0.000000}%
\psset{linecolor=dialinecolor}
\psline(41.695022,13.492628)(41.750278,13.602500)(41.805022,13.492373)
\newrgbcolor{dialinecolor}{1.000000 1.000000 1.000000}%
\psset{linecolor=dialinecolor}
\pspolygon*(45.246720,3.962500)(48.469522,3.962500)(47.791628,5.825000)(44.568825,5.825000)
\psset{linewidth=0.030000cm}
\psset{linestyle=solid}
\psset{linestyle=solid}
\setlinejoinmode{0}
\newrgbcolor{dialinecolor}{0.000000 0.000000 0.000000}%
\psset{linecolor=dialinecolor}
\pspolygon(45.246720,3.962500)(48.469522,3.962500)(47.791628,5.825000)(44.568825,5.825000)
\setfont{Helvetica}{0.529167}
\newrgbcolor{dialinecolor}{0.000000 0.000000 0.000000}%
\psset{linecolor=dialinecolor}
\rput(46.519174,4.492500){\psscalebox{1 -1}{Task-Trained}}
\setfont{Helvetica}{0.529167}
\newrgbcolor{dialinecolor}{0.000000 0.000000 0.000000}%
\psset{linecolor=dialinecolor}
\rput(46.519174,5.021667){\psscalebox{1 -1}{Embeddings}}
\setfont{Helvetica}{0.529167}
\newrgbcolor{dialinecolor}{0.000000 0.000000 0.000000}%
\psset{linecolor=dialinecolor}
\rput(46.519174,5.550833){\psscalebox{1 -1}{}}
\setfont{Helvetica}{0.776111}
\newrgbcolor{dialinecolor}{0.000000 0.000000 0.000000}%
\psset{linecolor=dialinecolor}
\rput[l](45.928825,5.522500){\psscalebox{1 -1}{$e_{i}^{t}$}}
\newrgbcolor{dialinecolor}{1.000000 1.000000 1.000000}%
\psset{linecolor=dialinecolor}
\pspolygon*(40.556325,2.362500)(40.556325,6.625000)(43.193825,6.625000)(43.193825,2.362500)
\psset{linewidth=0.030000cm}
\psset{linestyle=solid}
\psset{linestyle=solid}
\setlinejoinmode{0}
\newrgbcolor{dialinecolor}{0.000000 0.000000 0.000000}%
\psset{linecolor=dialinecolor}
\pspolygon(40.556325,2.362500)(40.556325,6.625000)(43.193825,6.625000)(43.193825,2.362500)
\setfont{Helvetica}{0.529167}
\newrgbcolor{dialinecolor}{0.000000 0.000000 0.000000}%
\psset{linecolor=dialinecolor}
\rput(41.875075,4.357083){\psscalebox{1 -1}{Parser}}
\setfont{Helvetica}{0.529167}
\newrgbcolor{dialinecolor}{0.000000 0.000000 0.000000}%
\psset{linecolor=dialinecolor}
\rput(41.875075,4.886250){\psscalebox{1 -1}{Training}}
\newrgbcolor{dialinecolor}{1.000000 1.000000 1.000000}%
\psset{linecolor=dialinecolor}
\pspolygon*(47.584220,12.687500)(50.807022,12.687500)(50.129128,14.550000)(46.906325,14.550000)
\psset{linewidth=0.030000cm}
\psset{linestyle=solid}
\psset{linestyle=solid}
\setlinejoinmode{0}
\newrgbcolor{dialinecolor}{0.000000 0.000000 0.000000}%
\psset{linecolor=dialinecolor}
\pspolygon(47.584220,12.687500)(50.807022,12.687500)(50.129128,14.550000)(46.906325,14.550000)
\setfont{Helvetica}{0.529167}
\newrgbcolor{dialinecolor}{0.000000 0.000000 0.000000}%
\psset{linecolor=dialinecolor}
\rput(48.856674,13.217500){\psscalebox{1 -1}{Mapped}}
\setfont{Helvetica}{0.529167}
\newrgbcolor{dialinecolor}{0.000000 0.000000 0.000000}%
\psset{linecolor=dialinecolor}
\rput(48.856674,13.746667){\psscalebox{1 -1}{Embeddings}}
\setfont{Helvetica}{0.529167}
\newrgbcolor{dialinecolor}{0.000000 0.000000 0.000000}%
\psset{linecolor=dialinecolor}
\rput(48.856674,14.275833){\psscalebox{1 -1}{}}
\psset{linewidth=0.030000cm}
\psset{linestyle=solid}
\psset{linestyle=solid}
\setlinecaps{0}
\newrgbcolor{dialinecolor}{0.000000 0.000000 0.000000}%
\psset{linecolor=dialinecolor}
\psline(45.018850,13.787447)(46.942784,13.787495)
\psset{linestyle=solid}
\setlinejoinmode{0}
\setlinecaps{0}
\newrgbcolor{dialinecolor}{0.000000 0.000000 0.000000}%
\psset{linecolor=dialinecolor}
\pspolygon*(47.100284,13.787499)(46.890281,13.892494)(46.942784,13.787495)(46.890287,13.682494)
\newrgbcolor{dialinecolor}{0.000000 0.000000 0.000000}%
\psset{linecolor=dialinecolor}
\pspolygon(47.100284,13.787499)(46.890281,13.892494)(46.942784,13.787495)(46.890287,13.682494)
\psset{linewidth=0.030000cm}
\psset{linestyle=solid}
\psset{linestyle=solid}
\setlinecaps{0}
\newrgbcolor{dialinecolor}{0.000000 0.000000 0.000000}%
\psset{linecolor=dialinecolor}
\psline(49.533825,12.625000)(49.533825,10.666041)
\psset{linestyle=solid}
\setlinejoinmode{0}
\setlinecaps{0}
\newrgbcolor{dialinecolor}{0.000000 0.000000 0.000000}%
\psset{linecolor=dialinecolor}
\pspolygon*(49.533825,10.508541)(49.638825,10.718541)(49.533825,10.666041)(49.428825,10.718541)
\newrgbcolor{dialinecolor}{0.000000 0.000000 0.000000}%
\psset{linecolor=dialinecolor}
\pspolygon(49.533825,10.508541)(49.638825,10.718541)(49.533825,10.666041)(49.428825,10.718541)
\setfont{Helvetica}{0.776111}
\newrgbcolor{dialinecolor}{0.000000 0.000000 0.000000}%
\psset{linecolor=dialinecolor}
\rput[l](48.216325,14.272500){\psscalebox{1 -1}{$e_{i}^{m}$}}
}\endpspicture

%% file: figs/parsing.tex
\ifx\setlinejoinmode\undefined
  \newcommand{\setlinejoinmode}[1]{}
\fi
\ifx\setlinecaps\undefined
  \newcommand{\setlinecaps}[1]{}
\fi
\ifx\setfont\undefined
  \newcommand{\setfont}[2]{}
\fi
\pspicture(33.385887,-33.476868)(49.080077,-16.706868)
\psscalebox{1.000000 -1.000000}{
\newrgbcolor{dialinecolor}{0.000000 0.000000 0.000000}%
\psset{linecolor=dialinecolor}
\newrgbcolor{diafillcolor}{1.000000 1.000000 1.000000}%
\psset{fillcolor=diafillcolor}
\psset{linewidth=0.030000cm}
\psset{linestyle=solid}
\psset{linestyle=solid}
\setlinejoinmode{0}
\setlinecaps{0}
\newrgbcolor{dialinecolor}{0.000000 0.000000 0.000000}%
\psset{linecolor=dialinecolor}
\psline(43.925077,21.199368)(46.583433,21.199368)(46.583433,27.393725)(43.043608,27.393725)
\psset{linestyle=solid}
\setlinejoinmode{0}
\setlinecaps{0}
\newrgbcolor{dialinecolor}{0.000000 0.000000 0.000000}%
\psset{linecolor=dialinecolor}
\pspolygon*(42.893608,27.393725)(43.093608,27.293725)(43.043608,27.393725)(43.093608,27.493725)
\newrgbcolor{dialinecolor}{0.000000 0.000000 0.000000}%
\psset{linecolor=dialinecolor}
\pspolygon(42.893608,27.393725)(43.093608,27.293725)(43.043608,27.393725)(43.093608,27.493725)
\psset{linewidth=0.030000cm}
\psset{linestyle=solid}
\psset{linestyle=solid}
\setlinecaps{0}
\newrgbcolor{dialinecolor}{0.000000 0.000000 0.000000}%
\psset{linecolor=dialinecolor}
\psline(40.975077,28.136868)(40.970184,28.821938)
\psset{linestyle=solid}
\setlinejoinmode{0}
\setlinecaps{0}
\newrgbcolor{dialinecolor}{0.000000 0.000000 0.000000}%
\psset{linecolor=dialinecolor}
\pspolygon*(40.969113,28.971934)(40.870544,28.771225)(40.970184,28.821938)(41.070539,28.772653)
\newrgbcolor{dialinecolor}{0.000000 0.000000 0.000000}%
\psset{linecolor=dialinecolor}
\pspolygon(40.969113,28.971934)(40.870544,28.771225)(40.970184,28.821938)(41.070539,28.772653)
\newrgbcolor{dialinecolor}{1.000000 1.000000 1.000000}%
\psset{linecolor=dialinecolor}
\pspolygon*(34.078782,19.936868)(37.301585,19.936868)(36.623690,21.799368)(33.400887,21.799368)
\psset{linewidth=0.030000cm}
\psset{linestyle=solid}
\psset{linestyle=solid}
\setlinejoinmode{0}
\newrgbcolor{dialinecolor}{0.000000 0.000000 0.000000}%
\psset{linecolor=dialinecolor}
\pspolygon(34.078782,19.936868)(37.301585,19.936868)(36.623690,21.799368)(33.400887,21.799368)
\setfont{Helvetica}{0.529167}
\newrgbcolor{dialinecolor}{0.000000 0.000000 0.000000}%
\psset{linecolor=dialinecolor}
\rput(35.351236,20.466868){\psscalebox{1 -1}{Initial}}
\setfont{Helvetica}{0.529167}
\newrgbcolor{dialinecolor}{0.000000 0.000000 0.000000}%
\psset{linecolor=dialinecolor}
\rput(35.351236,20.996035){\psscalebox{1 -1}{Embeddings}}
\setfont{Helvetica}{0.529167}
\newrgbcolor{dialinecolor}{0.000000 0.000000 0.000000}%
\psset{linecolor=dialinecolor}
\rput(35.351236,21.525202){\psscalebox{1 -1}{}}
\setfont{Helvetica}{0.776111}
\newrgbcolor{dialinecolor}{0.000000 0.000000 0.000000}%
\psset{linecolor=dialinecolor}
\rput[l](34.813736,21.530618){\psscalebox{1 -1}{$e_{i}^{o}$}}
\setfont{Helvetica}{0.800000}
\newrgbcolor{dialinecolor}{0.000000 0.000000 0.000000}%
\psset{linecolor=dialinecolor}
\rput[l](45.478827,31.129879){\psscalebox{1 -1}{}}
\psset{linewidth=0.030000cm}
\psset{linestyle=solid}
\psset{linestyle=solid}
\setlinecaps{0}
\newrgbcolor{dialinecolor}{0.000000 0.000000 0.000000}%
\psset{linecolor=dialinecolor}
\psline(43.712577,31.086868)(42.583610,31.097620)
\psset{linestyle=solid}
\setlinejoinmode{0}
\setlinecaps{0}
\newrgbcolor{dialinecolor}{0.000000 0.000000 0.000000}%
\psset{linecolor=dialinecolor}
\pspolygon*(42.433617,31.099049)(42.632655,30.997149)(42.583610,31.097620)(42.634560,31.197140)
\newrgbcolor{dialinecolor}{0.000000 0.000000 0.000000}%
\psset{linecolor=dialinecolor}
\pspolygon(42.433617,31.099049)(42.632655,30.997149)(42.583610,31.097620)(42.634560,31.197140)
\newrgbcolor{dialinecolor}{1.000000 1.000000 1.000000}%
\psset{linecolor=dialinecolor}
\pspolygon*(35.466189,29.361868)(38.081391,29.361868)(37.612779,30.649368)(34.997577,30.649368)
\psset{linewidth=0.030000cm}
\psset{linestyle=solid}
\psset{linestyle=solid}
\setlinejoinmode{0}
\newrgbcolor{dialinecolor}{0.000000 0.000000 0.000000}%
\psset{linecolor=dialinecolor}
\pspolygon(35.466189,29.361868)(38.081391,29.361868)(37.612779,30.649368)(34.997577,30.649368)
\setfont{Helvetica}{0.529167}
\newrgbcolor{dialinecolor}{0.000000 0.000000 0.000000}%
\psset{linecolor=dialinecolor}
\rput(36.539484,29.868952){\psscalebox{1 -1}{Testing }}
\setfont{Helvetica}{0.529167}
\newrgbcolor{dialinecolor}{0.000000 0.000000 0.000000}%
\psset{linecolor=dialinecolor}
\rput(36.539484,30.398118){\psscalebox{1 -1}{Sentences}}
\psset{linewidth=0.030000cm}
\psset{linestyle=solid}
\psset{linestyle=solid}
\setlinecaps{0}
\newrgbcolor{dialinecolor}{0.000000 0.000000 0.000000}%
\psset{linecolor=dialinecolor}
\psline(37.910333,30.141234)(39.259048,30.126461)
\psset{linestyle=solid}
\setlinejoinmode{0}
\setlinecaps{0}
\newrgbcolor{dialinecolor}{0.000000 0.000000 0.000000}%
\psset{linecolor=dialinecolor}
\pspolygon*(39.416538,30.124736)(39.207701,30.232029)(39.259048,30.126461)(39.205401,30.022042)
\newrgbcolor{dialinecolor}{0.000000 0.000000 0.000000}%
\psset{linecolor=dialinecolor}
\pspolygon(39.416538,30.124736)(39.207701,30.232029)(39.259048,30.126461)(39.205401,30.022042)
\psset{linewidth=0.030000cm}
\psset{linestyle=solid}
\psset{linestyle=solid}
\setlinecaps{0}
\setlinejoinmode{0}
\psset{linewidth=0.030000cm}
\setlinecaps{0}
\setlinejoinmode{0}
\psset{linestyle=solid}
\newrgbcolor{dialinecolor}{1.000000 1.000000 1.000000}%
\psset{linecolor=dialinecolor}
\pscustom{
\newpath
\moveto(35.406952,31.621868)
\lineto(38.466327,31.621868)
\curveto(38.007421,31.945368)(37.854452,32.107118)(37.854452,32.430618)
\curveto(37.854452,32.754118)(38.007421,32.915868)(38.466327,33.239368)
\lineto(35.406952,33.239368)
\curveto(34.948046,32.915868)(34.795077,32.754118)(34.795077,32.430618)
\curveto(34.795077,32.107118)(34.948046,31.945368)(35.406952,31.621868)
\fill[fillstyle=solid,fillcolor=diafillcolor,linecolor=diafillcolor]}
\newrgbcolor{dialinecolor}{0.000000 0.000000 0.000000}%
\psset{linecolor=dialinecolor}
\pscustom{
\newpath
\moveto(35.406952,31.621868)
\lineto(38.466327,31.621868)
\curveto(38.007421,31.945368)(37.854452,32.107118)(37.854452,32.430618)
\curveto(37.854452,32.754118)(38.007421,32.915868)(38.466327,33.239368)
\lineto(35.406952,33.239368)
\curveto(34.948046,32.915868)(34.795077,32.754118)(34.795077,32.430618)
\curveto(34.795077,32.107118)(34.948046,31.945368)(35.406952,31.621868)
\stroke}
\setfont{Helvetica}{0.529167}
\newrgbcolor{dialinecolor}{0.000000 0.000000 0.000000}%
\psset{linecolor=dialinecolor}
\rput(36.630702,32.033743){\psscalebox{1 -1}{Model}}
\setfont{Helvetica}{0.529167}
\newrgbcolor{dialinecolor}{0.000000 0.000000 0.000000}%
\psset{linecolor=dialinecolor}
\rput(36.630702,32.562910){\psscalebox{1 -1}{Parameters}}
\setfont{Helvetica}{0.529167}
\newrgbcolor{dialinecolor}{0.000000 0.000000 0.000000}%
\psset{linecolor=dialinecolor}
\rput(36.630702,33.092077){\psscalebox{1 -1}{}}
\psset{linewidth=0.030000cm}
\psset{linestyle=solid}
\psset{linestyle=solid}
\setlinecaps{0}
\newrgbcolor{dialinecolor}{0.000000 0.000000 0.000000}%
\psset{linecolor=dialinecolor}
\psline(37.947833,32.416234)(39.196549,32.401606)
\psset{linestyle=solid}
\setlinejoinmode{0}
\setlinecaps{0}
\newrgbcolor{dialinecolor}{0.000000 0.000000 0.000000}%
\psset{linecolor=dialinecolor}
\pspolygon*(39.354038,32.399761)(39.145283,32.507214)(39.196549,32.401606)(39.142823,32.297228)
\newrgbcolor{dialinecolor}{0.000000 0.000000 0.000000}%
\psset{linecolor=dialinecolor}
\pspolygon(39.354038,32.399761)(39.145283,32.507214)(39.196549,32.401606)(39.142823,32.297228)
\setfont{Helvetica}{0.705556}
\newrgbcolor{dialinecolor}{0.000000 0.000000 0.000000}%
\psset{linecolor=dialinecolor}
\rput[l](36.020077,33.034368){\psscalebox{1 -1}{$W$}}
\psset{linewidth=0.030000cm}
\psset{linestyle=solid}
\psset{linestyle=solid}
\setlinecaps{0}
\newrgbcolor{dialinecolor}{0.000000 0.000000 0.000000}%
\psset{linecolor=dialinecolor}
\psline(37.086281,21.005096)(38.202733,20.985846)
\psset{linestyle=solid}
\setlinejoinmode{0}
\setlinecaps{0}
\newrgbcolor{dialinecolor}{0.000000 0.000000 0.000000}%
\psset{linecolor=dialinecolor}
\pspolygon*(38.360209,20.983131)(38.152051,21.091736)(38.202733,20.985846)(38.148430,20.881767)
\newrgbcolor{dialinecolor}{0.000000 0.000000 0.000000}%
\psset{linecolor=dialinecolor}
\pspolygon(38.360209,20.983131)(38.152051,21.091736)(38.202733,20.985846)(38.148430,20.881767)
\newrgbcolor{dialinecolor}{1.000000 1.000000 1.000000}%
\psset{linecolor=dialinecolor}
\pspolygon*(38.575077,18.890000)(38.575077,23.926667)(39.090833,23.926667)(39.090833,18.890000)
\psset{linewidth=0.030000cm}
\psset{linestyle=solid}
\psset{linestyle=solid}
\setlinejoinmode{0}
\newrgbcolor{dialinecolor}{0.000000 0.000000 0.000000}%
\psset{linecolor=dialinecolor}
\pspolygon(38.575077,18.890000)(38.575077,23.926667)(39.090833,23.926667)(39.090833,18.890000)
\setfont{Helvetica}{0.800000}
\newrgbcolor{dialinecolor}{0.000000 0.000000 0.000000}%
\psset{linecolor=dialinecolor}
\rput(38.832955,21.603333){\psscalebox{1 -1}{}}
\psset{linewidth=0.030000cm}
\psset{linestyle=solid}
\psset{linestyle=solid}
\setlinecaps{0}
\newrgbcolor{dialinecolor}{0.000000 0.000000 0.000000}%
\psset{linecolor=dialinecolor}
\psline(39.125077,20.061868)(39.729059,20.071455)
\psset{linestyle=solid}
\setlinejoinmode{0}
\setlinecaps{0}
\newrgbcolor{dialinecolor}{0.000000 0.000000 0.000000}%
\psset{linecolor=dialinecolor}
\pspolygon*(39.879040,20.073836)(39.677478,20.170649)(39.729059,20.071455)(39.680653,19.970674)
\newrgbcolor{dialinecolor}{0.000000 0.000000 0.000000}%
\psset{linecolor=dialinecolor}
\pspolygon(39.879040,20.073836)(39.677478,20.170649)(39.729059,20.071455)(39.680653,19.970674)
\newrgbcolor{dialinecolor}{1.000000 1.000000 1.000000}%
\psset{linecolor=dialinecolor}
\pspolygon*(43.385077,18.886868)(43.385077,23.923535)(43.900833,23.923535)(43.900833,18.886868)
\psset{linewidth=0.030000cm}
\psset{linestyle=solid}
\psset{linestyle=solid}
\setlinejoinmode{0}
\newrgbcolor{dialinecolor}{0.000000 0.000000 0.000000}%
\psset{linecolor=dialinecolor}
\pspolygon(43.385077,18.886868)(43.385077,23.923535)(43.900833,23.923535)(43.900833,18.886868)
\setfont{Helvetica}{0.800000}
\newrgbcolor{dialinecolor}{0.000000 0.000000 0.000000}%
\psset{linecolor=dialinecolor}
\rput(43.642955,21.600202){\psscalebox{1 -1}{}}
\psset{linewidth=0.030000cm}
\psset{linestyle=solid}
\psset{linestyle=solid}
\setlinecaps{0}
\newrgbcolor{dialinecolor}{0.000000 0.000000 0.000000}%
\psset{linecolor=dialinecolor}
\psline(39.135313,20.959099)(39.739295,20.968686)
\psset{linestyle=solid}
\setlinejoinmode{0}
\setlinecaps{0}
\newrgbcolor{dialinecolor}{0.000000 0.000000 0.000000}%
\psset{linecolor=dialinecolor}
\pspolygon*(39.889277,20.971067)(39.687715,21.067880)(39.739295,20.968686)(39.690889,20.867905)
\newrgbcolor{dialinecolor}{0.000000 0.000000 0.000000}%
\psset{linecolor=dialinecolor}
\pspolygon(39.889277,20.971067)(39.687715,21.067880)(39.739295,20.968686)(39.690889,20.867905)
\psset{linewidth=0.030000cm}
\psset{linestyle=solid}
\psset{linestyle=solid}
\setlinecaps{0}
\newrgbcolor{dialinecolor}{0.000000 0.000000 0.000000}%
\psset{linecolor=dialinecolor}
\psline(39.110313,21.734099)(39.714295,21.743686)
\psset{linestyle=solid}
\setlinejoinmode{0}
\setlinecaps{0}
\newrgbcolor{dialinecolor}{0.000000 0.000000 0.000000}%
\psset{linecolor=dialinecolor}
\pspolygon*(39.864277,21.746067)(39.662715,21.842880)(39.714295,21.743686)(39.665889,21.642905)
\newrgbcolor{dialinecolor}{0.000000 0.000000 0.000000}%
\psset{linecolor=dialinecolor}
\pspolygon(39.864277,21.746067)(39.662715,21.842880)(39.714295,21.743686)(39.665889,21.642905)
\psset{linewidth=0.030000cm}
\psset{linestyle=solid}
\psset{linestyle=solid}
\setlinecaps{0}
\newrgbcolor{dialinecolor}{0.000000 0.000000 0.000000}%
\psset{linecolor=dialinecolor}
\psline(39.130313,22.544099)(39.734295,22.553686)
\psset{linestyle=solid}
\setlinejoinmode{0}
\setlinecaps{0}
\newrgbcolor{dialinecolor}{0.000000 0.000000 0.000000}%
\psset{linecolor=dialinecolor}
\pspolygon*(39.884277,22.556067)(39.682715,22.652880)(39.734295,22.553686)(39.685889,22.452905)
\newrgbcolor{dialinecolor}{0.000000 0.000000 0.000000}%
\psset{linecolor=dialinecolor}
\pspolygon(39.884277,22.556067)(39.682715,22.652880)(39.734295,22.553686)(39.685889,22.452905)
\psset{linewidth=0.030000cm}
\psset{linestyle=solid}
\psset{linestyle=solid}
\setlinecaps{0}
\newrgbcolor{dialinecolor}{0.000000 0.000000 0.000000}%
\psset{linecolor=dialinecolor}
\psline(39.135313,23.359099)(39.739295,23.368686)
\psset{linestyle=solid}
\setlinejoinmode{0}
\setlinecaps{0}
\newrgbcolor{dialinecolor}{0.000000 0.000000 0.000000}%
\psset{linecolor=dialinecolor}
\pspolygon*(39.889277,23.371067)(39.687715,23.467880)(39.739295,23.368686)(39.690889,23.267905)
\newrgbcolor{dialinecolor}{0.000000 0.000000 0.000000}%
\psset{linecolor=dialinecolor}
\pspolygon(39.889277,23.371067)(39.687715,23.467880)(39.739295,23.368686)(39.690889,23.267905)
\psset{linewidth=0.030000cm}
\psset{linestyle=solid}
\psset{linestyle=solid}
\setlinecaps{0}
\newrgbcolor{dialinecolor}{0.000000 0.000000 0.000000}%
\psset{linecolor=dialinecolor}
\psline(42.522813,20.046599)(43.126795,20.056186)
\psset{linestyle=solid}
\setlinejoinmode{0}
\setlinecaps{0}
\newrgbcolor{dialinecolor}{0.000000 0.000000 0.000000}%
\psset{linecolor=dialinecolor}
\pspolygon*(43.276777,20.058567)(43.075215,20.155380)(43.126795,20.056186)(43.078389,19.955405)
\newrgbcolor{dialinecolor}{0.000000 0.000000 0.000000}%
\psset{linecolor=dialinecolor}
\pspolygon(43.276777,20.058567)(43.075215,20.155380)(43.126795,20.056186)(43.078389,19.955405)
\psset{linewidth=0.030000cm}
\psset{linestyle=solid}
\psset{linestyle=solid}
\setlinecaps{0}
\newrgbcolor{dialinecolor}{0.000000 0.000000 0.000000}%
\psset{linecolor=dialinecolor}
\psline(42.480313,20.931599)(43.084295,20.941186)
\psset{linestyle=solid}
\setlinejoinmode{0}
\setlinecaps{0}
\newrgbcolor{dialinecolor}{0.000000 0.000000 0.000000}%
\psset{linecolor=dialinecolor}
\pspolygon*(43.234277,20.943567)(43.032715,21.040380)(43.084295,20.941186)(43.035889,20.840405)
\newrgbcolor{dialinecolor}{0.000000 0.000000 0.000000}%
\psset{linecolor=dialinecolor}
\pspolygon(43.234277,20.943567)(43.032715,21.040380)(43.084295,20.941186)(43.035889,20.840405)
\psset{linewidth=0.030000cm}
\psset{linestyle=solid}
\psset{linestyle=solid}
\setlinecaps{0}
\newrgbcolor{dialinecolor}{0.000000 0.000000 0.000000}%
\psset{linecolor=dialinecolor}
\psline(42.487813,21.704099)(43.091795,21.713686)
\psset{linestyle=solid}
\setlinejoinmode{0}
\setlinecaps{0}
\newrgbcolor{dialinecolor}{0.000000 0.000000 0.000000}%
\psset{linecolor=dialinecolor}
\pspolygon*(43.241777,21.716067)(43.040215,21.812880)(43.091795,21.713686)(43.043389,21.612905)
\newrgbcolor{dialinecolor}{0.000000 0.000000 0.000000}%
\psset{linecolor=dialinecolor}
\pspolygon(43.241777,21.716067)(43.040215,21.812880)(43.091795,21.713686)(43.043389,21.612905)
\psset{linewidth=0.030000cm}
\psset{linestyle=solid}
\psset{linestyle=solid}
\setlinecaps{0}
\newrgbcolor{dialinecolor}{0.000000 0.000000 0.000000}%
\psset{linecolor=dialinecolor}
\psline(42.495313,22.501599)(43.099295,22.511186)
\psset{linestyle=solid}
\setlinejoinmode{0}
\setlinecaps{0}
\newrgbcolor{dialinecolor}{0.000000 0.000000 0.000000}%
\psset{linecolor=dialinecolor}
\pspolygon*(43.249277,22.513567)(43.047715,22.610380)(43.099295,22.511186)(43.050889,22.410405)
\newrgbcolor{dialinecolor}{0.000000 0.000000 0.000000}%
\psset{linecolor=dialinecolor}
\pspolygon(43.249277,22.513567)(43.047715,22.610380)(43.099295,22.511186)(43.050889,22.410405)
\psset{linewidth=0.030000cm}
\psset{linestyle=solid}
\psset{linestyle=solid}
\setlinecaps{0}
\newrgbcolor{dialinecolor}{0.000000 0.000000 0.000000}%
\psset{linecolor=dialinecolor}
\psline(42.515313,23.349099)(43.119295,23.358686)
\psset{linestyle=solid}
\setlinejoinmode{0}
\setlinecaps{0}
\newrgbcolor{dialinecolor}{0.000000 0.000000 0.000000}%
\psset{linecolor=dialinecolor}
\pspolygon*(43.269277,23.361067)(43.067715,23.457880)(43.119295,23.358686)(43.070889,23.257905)
\newrgbcolor{dialinecolor}{0.000000 0.000000 0.000000}%
\psset{linecolor=dialinecolor}
\pspolygon(43.269277,23.361067)(43.067715,23.457880)(43.119295,23.358686)(43.070889,23.257905)
\newrgbcolor{dialinecolor}{1.000000 1.000000 1.000000}%
\psset{linecolor=dialinecolor}
\pspolygon*(39.997577,17.911868)(39.997577,24.711868)(42.462994,24.711868)(42.462994,17.911868)
\psset{linewidth=0.030000cm}
\psset{linestyle=solid}
\psset{linestyle=solid}
\setlinejoinmode{0}
\newrgbcolor{dialinecolor}{0.000000 0.000000 0.000000}%
\psset{linecolor=dialinecolor}
\pspolygon(39.997577,17.911868)(39.997577,24.711868)(42.462994,24.711868)(42.462994,17.911868)
\setfont{Helvetica}{0.531989}
\newrgbcolor{dialinecolor}{0.000000 0.000000 0.000000}%
\psset{linecolor=dialinecolor}
\rput(41.230285,21.174879){\psscalebox{1 -1}{Non-Linear}}
\setfont{Helvetica}{0.531989}
\newrgbcolor{dialinecolor}{0.000000 0.000000 0.000000}%
\psset{linecolor=dialinecolor}
\rput(41.230285,21.706868){\psscalebox{1 -1}{Layer}}
\psset{linewidth=0.010000cm}
\psset{linestyle=dashed,dash=1.000000 1.000000}
\psset{linestyle=dashed,dash=1.000000 1.000000}
\setlinecaps{0}
\newrgbcolor{dialinecolor}{0.000000 0.000000 0.000000}%
\psset{linecolor=dialinecolor}
\psline(37.950077,16.761868)(37.937577,25.099368)
\psset{linewidth=0.010000cm}
\psset{linestyle=dashed,dash=1.000000 1.000000}
\psset{linestyle=dashed,dash=1.000000 1.000000}
\setlinecaps{0}
\newrgbcolor{dialinecolor}{0.000000 0.000000 0.000000}%
\psset{linecolor=dialinecolor}
\psline(44.987577,16.711868)(44.987577,25.099368)
\psset{linewidth=0.010000cm}
\psset{linestyle=dashed,dash=1.000000 1.000000}
\psset{linestyle=dashed,dash=1.000000 1.000000}
\setlinecaps{0}
\newrgbcolor{dialinecolor}{0.000000 0.000000 0.000000}%
\psset{linecolor=dialinecolor}
\psline(37.962577,16.749368)(44.975077,16.736868)
\psset{linewidth=0.010000cm}
\psset{linestyle=dashed,dash=1.000000 1.000000}
\psset{linestyle=dashed,dash=1.000000 1.000000}
\setlinecaps{0}
\newrgbcolor{dialinecolor}{0.000000 0.000000 0.000000}%
\psset{linecolor=dialinecolor}
\psline(37.950086,25.151877)(44.962586,25.139377)
\setfont{Helvetica}{0.705556}
\newrgbcolor{dialinecolor}{0.000000 0.000000 0.000000}%
\psset{linecolor=dialinecolor}
\rput[l](39.800077,17.474368){\psscalebox{1 -1}{Mapper Function}}
\psset{linewidth=0.030000cm}
\psset{linestyle=solid}
\psset{linestyle=solid}
\setlinejoinmode{0}
\setlinecaps{0}
\newrgbcolor{dialinecolor}{0.000000 0.000000 0.000000}%
\psset{linecolor=dialinecolor}
\pscustom{
\newpath
\moveto(40.250077,20.499368)
\curveto(41.175527,20.499368)(41.324627,19.074368)(42.250077,19.074368)
\stroke}
\psset{linewidth=0.000100cm}
\psset{linestyle=solid}
\psset{linestyle=solid}
\setlinecaps{0}
\newrgbcolor{dialinecolor}{0.000000 0.000000 0.000000}%
\psset{linecolor=dialinecolor}
\psline(40.425301,19.824368)(42.037354,19.824368)
\psset{linewidth=0.000100cm}
\psset{linestyle=solid}
\setlinejoinmode{0}
\setlinecaps{0}
\newrgbcolor{dialinecolor}{0.000000 0.000000 0.000000}%
\psset{linecolor=dialinecolor}
\psline(40.535189,19.769368)(40.425189,19.824368)(40.535189,19.879368)
\psset{linewidth=0.000100cm}
\psset{linestyle=solid}
\setlinejoinmode{0}
\setlinecaps{0}
\newrgbcolor{dialinecolor}{0.000000 0.000000 0.000000}%
\psset{linecolor=dialinecolor}
\psline(41.927465,19.879368)(42.037465,19.824368)(41.927465,19.769368)
\psset{linewidth=0.000000cm}
\psset{linestyle=solid}
\psset{linestyle=solid}
\setlinecaps{0}
\newrgbcolor{dialinecolor}{0.000000 0.000000 0.000000}%
\psset{linecolor=dialinecolor}
\psline(41.207577,18.646868)(41.212577,20.799368)
\psset{linewidth=0.000000cm}
\psset{linestyle=solid}
\setlinejoinmode{0}
\setlinecaps{0}
\newrgbcolor{dialinecolor}{0.000000 0.000000 0.000000}%
\psset{linecolor=dialinecolor}
\psline(41.262833,18.756740)(41.207577,18.646868)(41.152833,18.756996)
\psset{linewidth=0.000000cm}
\psset{linestyle=solid}
\setlinejoinmode{0}
\setlinecaps{0}
\newrgbcolor{dialinecolor}{0.000000 0.000000 0.000000}%
\psset{linecolor=dialinecolor}
\psline(41.157322,20.689496)(41.212577,20.799368)(41.267321,20.689241)
\newrgbcolor{dialinecolor}{1.000000 1.000000 1.000000}%
\psset{linecolor=dialinecolor}
\pspolygon*(39.743385,25.919785)(43.158789,25.919785)(42.377466,28.066452)(38.962063,28.066452)
\psset{linewidth=0.030000cm}
\psset{linestyle=solid}
\psset{linestyle=solid}
\setlinejoinmode{0}
\newrgbcolor{dialinecolor}{0.000000 0.000000 0.000000}%
\psset{linecolor=dialinecolor}
\pspolygon(39.743385,25.919785)(43.158789,25.919785)(42.377466,28.066452)(38.962063,28.066452)
\setfont{Helvetica}{0.529167}
\newrgbcolor{dialinecolor}{0.000000 0.000000 0.000000}%
\psset{linecolor=dialinecolor}
\rput(41.060426,26.327285){\psscalebox{1 -1}{Unseen}}
\setfont{Helvetica}{0.529167}
\newrgbcolor{dialinecolor}{0.000000 0.000000 0.000000}%
\psset{linecolor=dialinecolor}
\rput(41.060426,26.856452){\psscalebox{1 -1}{Mapped}}
\setfont{Helvetica}{0.529167}
\newrgbcolor{dialinecolor}{0.000000 0.000000 0.000000}%
\psset{linecolor=dialinecolor}
\rput(41.060426,27.385618){\psscalebox{1 -1}{Embeddings}}
\setfont{Helvetica}{0.529167}
\newrgbcolor{dialinecolor}{0.000000 0.000000 0.000000}%
\psset{linecolor=dialinecolor}
\rput(41.060426,27.914785){\psscalebox{1 -1}{}}
\setfont{Helvetica}{0.776111}
\newrgbcolor{dialinecolor}{0.000000 0.000000 0.000000}%
\psset{linecolor=dialinecolor}
    \rput[l](40.507577,27.859368){\psscalebox{1 -1}{$e_{i}^{m}$}}
\newrgbcolor{dialinecolor}{1.000000 1.000000 1.000000}%
\psset{linecolor=dialinecolor}
\pspolygon*(44.391400,30.036868)(47.806804,30.036868)(47.025481,32.183535)(43.610077,32.183535)
\psset{linewidth=0.030000cm}
\psset{linestyle=solid}
\psset{linestyle=solid}
\setlinejoinmode{0}
\newrgbcolor{dialinecolor}{0.000000 0.000000 0.000000}%
\psset{linecolor=dialinecolor}
\pspolygon(44.391400,30.036868)(47.806804,30.036868)(47.025481,32.183535)(43.610077,32.183535)
\setfont{Helvetica}{0.529167}
\newrgbcolor{dialinecolor}{0.000000 0.000000 0.000000}%
\psset{linecolor=dialinecolor}
\rput(45.708440,30.444368){\psscalebox{1 -1}{Seen}}
\setfont{Helvetica}{0.529167}
\newrgbcolor{dialinecolor}{0.000000 0.000000 0.000000}%
\psset{linecolor=dialinecolor}
\rput(45.708440,30.973535){\psscalebox{1 -1}{Task-Trained}}
\setfont{Helvetica}{0.529167}
\newrgbcolor{dialinecolor}{0.000000 0.000000 0.000000}%
\psset{linecolor=dialinecolor}
\rput(45.708440,31.502702){\psscalebox{1 -1}{Embeddings}}
\setfont{Helvetica}{0.529167}
\newrgbcolor{dialinecolor}{0.000000 0.000000 0.000000}%
\psset{linecolor=dialinecolor}
\rput(45.708440,32.031868){\psscalebox{1 -1}{}}
\setfont{Helvetica}{0.776111}
\newrgbcolor{dialinecolor}{0.000000 0.000000 0.000000}%
\psset{linecolor=dialinecolor}
\rput[l](45.082577,31.921868){\psscalebox{1 -1}{$e_{i}^{t}$}}
\newrgbcolor{dialinecolor}{1.000000 1.000000 1.000000}%
\psset{linecolor=dialinecolor}
\pspolygon*(39.675077,29.199368)(39.675077,33.461868)(42.312577,33.461868)(42.312577,29.199368)
\psset{linewidth=0.030000cm}
\psset{linestyle=solid}
\psset{linestyle=solid}
\setlinejoinmode{0}
\newrgbcolor{dialinecolor}{0.000000 0.000000 0.000000}%
\psset{linecolor=dialinecolor}
\pspolygon(39.675077,29.199368)(39.675077,33.461868)(42.312577,33.461868)(42.312577,29.199368)
\setfont{Helvetica}{0.529167}
\newrgbcolor{dialinecolor}{0.000000 0.000000 0.000000}%
\psset{linecolor=dialinecolor}
\rput(40.993827,31.458535){\psscalebox{1 -1}{Parser}}
}\endpspicture